\documentclass[lettersize,journal]{IEEEtran}
\usepackage{amsmath,amsfonts}
\usepackage{algorithmic}
\usepackage{algorithm}
\usepackage{array}
\usepackage[caption=false,font=normalsize,labelfont=sf,textfont=sf]{subfig}
\usepackage{textcomp}
\usepackage{stfloats}
\usepackage{url}
\usepackage{verbatim}
\usepackage{graphicx}
\usepackage{cite}
\usepackage{multirow}
\usepackage{booktabs}

\usepackage[square,sort&compress,numbers]{natbib}
\usepackage{textcomp}
\usepackage{supertabular}
\usepackage{caption}

\usepackage{color, xcolor}
\usepackage{soul}
\soulregister\cite7 
\soulregister\ref7
\soulregister\subsection7

\begin{document}

\title{Object Detection in 20 Years: A Survey}

\author{Zhengxia~Zou$^\star$, Keyan~Chen, Zhenwei~Shi,~\IEEEmembership{Member,~IEEE}, Yuhong Guo, and Jieping Ye$^\star$,~\IEEEmembership{Fellow,~IEEE}
\thanks{The work was supported by the National Natural Science Foundation of China under Grant 62125102, the National Key Research and Development Program of China (Titled ``Brain-inspired General Vision Models and Applications''), and the Fundamental Research Funds for the Central Universities. \emph{(Corresponding Author: Zhengxia Zou (zhengxiazou@buaa.edu.cn) and Jieping Ye (jpye@umich.edu))}.}

\thanks{Zhengxia Zou is with the Department of Guidance, Navigation and Control, School of Astronautics, Beihang University, Beijing 100191, China, and also with Shanghai Artificial Intelligence Laboratory, Shanghai 200232, China.}

\thanks{Keyan Chen and Zhenwei Shi are with the Image Processing Center, School of Astronautics, and with the Beijing Key Laboratory of Digital Media, and with the State Key Laboratory of Virtual Reality Technology and Systems, Beihang University, Beijing 100191, China, and also with the Shanghai Artificial Intelligence Laboratory, Shanghai 200232, China.}

\thanks{Yuhong Guo is with the School of Computer Science, Carleton University, Ottawa, Ontario, K1S 5B6, Canada.}

\thanks{Jieping Ye is with the Alibaba Group, Hangzhou 310030, China.}
}

\maketitle

\begin{abstract}
Object detection, as of one the most fundamental and challenging problems in computer vision, has received great attention in recent years. Over the past two decades, we have seen a rapid technological evolution of object detection and its profound impact on the entire computer vision field. If we consider today's object detection technique as a revolution driven by deep learning, then back in the 1990s, we would see the ingenious thinking and long-term perspective design of early computer vision. This paper extensively reviews this fast-moving research field in the light of technical evolution, spanning over a quarter-century's time (from the 1990s to 2022). A number of topics have been covered in this paper, including the milestone detectors in history, detection datasets, metrics, fundamental building blocks of the detection system, speed-up techniques, and the recent state-of-the-art detection methods.
\end{abstract}


\section{Introduction}\label{sec:introduction}

\IEEEPARstart{O}{bject} detection is an important computer vision task that deals with detecting instances of visual objects of a certain class (such as humans, animals, or cars) in digital images. The goal of object detection is to develop computational models and techniques that provide one of the most basic pieces of knowledge needed by computer vision applications: \textit{What objects are where?} The two most significant metrics for object detection are accuracy (including classification accuracy and localization accuracy) and speed.

\begin{figure}
  \centering{\includegraphics[width=\linewidth]{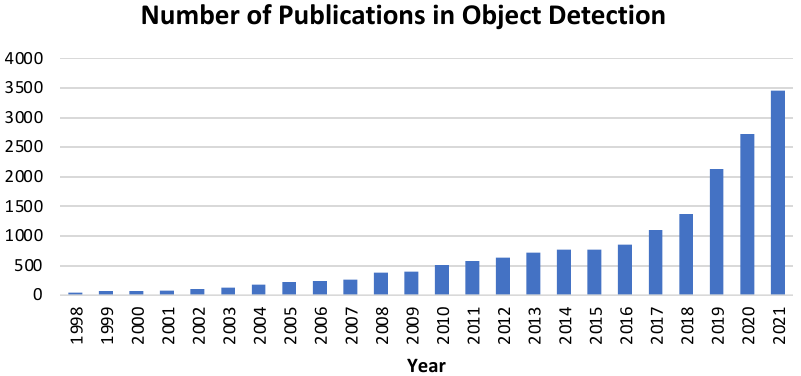}}\\
  \caption{The increasing number of publications in object detection from 1998 to 2021. (Data from Google scholar advanced search: \textit{allintitle: ``object detection'' OR ``detecting objects''}.)} \label{figure:number-of-papers}
\end{figure}

Object detection serves as a basis for many other computer vision tasks, such as instance segmentation \cite{ECCV14-SimultDetecSeg, hariharan2015hypercolumns, CVPR16-InstAwareSeg, ICCV17-MaskRCNN}, image captioning \cite{CVPR15-VisSemAlign, ICML15-ShowAttendTell, TPAMI18-CaptVQA}, object tracking \cite{kang2018t}, etc. In recent years, the rapid development of deep learning techniques \cite{Nature15-SurveyDL}  has greatly promoted the progress of object detection, leading to remarkable breakthroughs and propelling it to a research hot-spot with unprecedented attention. Object detection has now been widely used in many real-world applications, such as autonomous driving, robot vision, video surveillance, etc. Fig.\ \ref{figure:number-of-papers} shows the growing number of publications that are associated with ``object detection'' over the past two decades. 

\begin{figure*} \centering{\includegraphics[width=\linewidth]{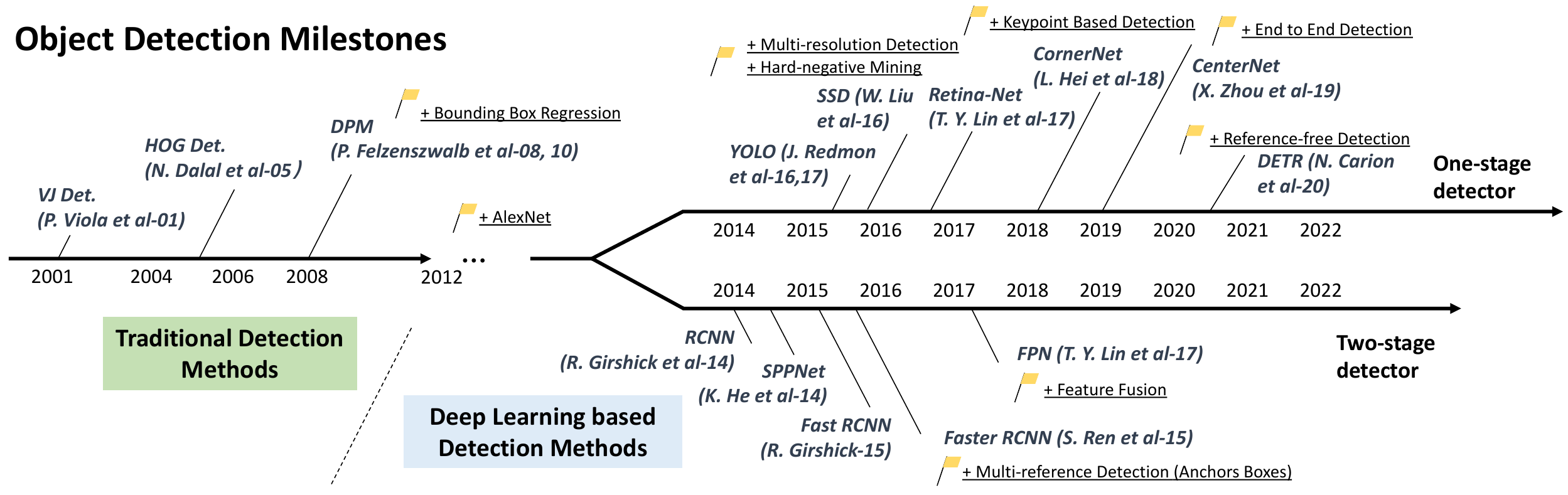}}\\
  \caption{ A road map of object detection. Milestone detectors in this figure: VJ Det.\ \cite{CVPR01-VJ,IJCV04-VJ}, HOG Det.\ \cite{CVPR05-HOG}, DPM \cite{CVPR08-DPM,CVPR10-DPM,TPAMI10-DPM}, RCNN \cite{CVPR14-RCNN}, SPPNet \cite{ECCV14-SPPNet}, Fast RCNN \cite{ICCV15-FastRCNN}, Faster RCNN \cite{NIPS15-FasterRCNN}, YOLO \cite{CVPR16-YOLO, redmon2018yolov3, bochkovskiy2020yolov4}, SSD \cite{ECCV16-SSD}, FPN \cite{CVPR17-FPN}, Retina-Net \cite{ICCV17-Focal}, CornerNet \cite{law2018cornernet}, CenterNet \cite{zhao2019object}, DETR \cite{carion2020end}.}\label{figure:mile-stones}
\end{figure*}

As different detection tasks have totally different objectives and constraints, their difficulties may vary from each other. In addition to some common challenges in other computer vision tasks such as objects under different viewpoints, illuminations, and intraclass variations, the challenges in object detection include but are not limited to the following aspects: object rotation and scale changes (e.g., small objects), accurate object localization, dense and occluded object detection, speed up of detection, etc. In Sec. \ref{sec:recent-advances}, we will give a more detailed analysis of these topics.

This survey seeks to provide novices with a complete grasp of object detection technology from many viewpoints, with an emphasis on its evolution. The key features are three-folds: A comprehensive review in the light of technical evolutions, an in-depth exploration of the key technologies and the recent state of the arts, and a comprehensive analysis of detection speed-up techniques. The main clue focuses on the past, present, and future, complemented with some other  necessary components in object detection, like datasets, metrics, and acceleration techniques. Standing on the technical highway, this survey aims to present the evolution of related technologies, allowing readers to grasp the essential concepts and find potential future directions, while neglecting their technical specifics.

The rest of this paper is organized as follows. In Section \ref{sec:20years}, we review the 20 years' evolution of object detection. In Section \ref{sec:speedup}, we review the speed-up techniques in object detection. The state-of-the-art detection methods of the recent three years are reviewed in Section \ref{sec:recent-advances}. In Section \ref{sec:conclusion}, we conclude this paper and make a deep analysis of the further research directions.

\section{Object Detection in 20 Years}\label{sec:20years}

In this section, we will review the history of object detection from multiple views, including milestone detectors, datasets, metrics and the evolution of key techniques.

\subsection{A Road Map of Object Detection}

In the past two decades, it is widely accepted that the progress of object detection has generally gone through two historical periods: ``traditional object detection period (before 2014)'' and ``deep learning based detection period (after 2014)'', as shown in Fig.\ \ref{figure:mile-stones}. In the following, we will summarize the milestone detectors of this period, with the emergence time and performance serving as the main clue to highlight the behind driving technology, seeing Fig. \ref{figure:accuracy-improvements}. 

\subsubsection{Milestones: Traditional Detectors}

If we consider today’s object detection technique as a revolution driven by deep learning, then back in the 1990s, we would see the ingenious design and long-term perspective of early computer vision. Most of the early object detection algorithms were built based on handcrafted features. Due to the lack of effective image representation at that time, people have to design sophisticated feature representations and a variety of speed-up skills.

\textbf{Viola Jones Detectors:}
In 2001, P. Viola and M. Jones achieved real-time detection of human faces for the first time without any constraints (e.g., skin color segmentation) \cite{CVPR01-VJ, IJCV04-VJ}. Running on a 700MHz Pentium III CPU, the detector was tens or even hundreds of times faster than other algorithms in its time under comparable detection accuracy. 
The VJ detector follows a most straightforward way of detection, i.e., sliding windows: to go through all possible locations and scales in an image to see if any window contains a human face. Although it seems to be a very simple process, the calculation behind it was far beyond the computer's power of its time. The VJ detector has dramatically improved its detection speed by incorporating three important techniques: ``integral image'', ``feature selection'', and ``detection cascades'' (to be introduced in section \ref{sec:speedup}).

\textbf{HOG Detector:}
In 2005, N. Dalal and B. Triggs proposed Histogram of Oriented Gradients (HOG) feature descriptor \cite{CVPR05-HOG}. HOG can be considered as an important improvement of the scale-invariant feature transform \cite{ICCV99-SIFT,IJCV04-SIFT} and shape contexts \cite{TPAMI02-ShapeContexts} of its time. To balance the feature invariance (including translation, scale, illumination, etc) and the nonlinearity, the HOG descriptor is designed to be computed on a dense grid of uniformly spaced cells and use overlapping local contrast normalization (on ``blocks''). Although HOG can be used to detect a variety of object classes, it was motivated primarily by the problem of pedestrian detection. To detect objects of different sizes, the HOG detector rescales the input image for multiple times while keeping the size of a detection window unchanged. The HOG detector has been an important foundation of many object detectors \cite{CVPR08-DPM,CVPR10-DPM,ICCV11-Exemplar} and a large variety of computer vision applications for many years.

\textbf{Deformable Part-based Model (DPM):}
DPM, as the winners of VOC-07, -08, and -09 detection challenges, was the epitome of the traditional object detection methods. DPM was originally proposed by P. Felzenszwalb \cite{CVPR08-DPM} in 2008 as an extension of the HOG detector.
It follows the detection philosophy of ``divide and conquer'', where the training can be simply considered as the learning of a proper way of decomposing an object, and the inference can be considered as an ensemble of detections on different object parts. For example, the problem of detecting a ``car'' can be decomposed to the detection of its window, body, and wheels. This part of the work, a.k.a.\ ``star-model'', was introduced by P. Felzenszwalb \emph{et al.} \cite{CVPR08-DPM}. Later on, R. Girshick has further extended the star model to the ``mixture models'' to deal with the objects in the real world under more significant variations and has made a series of other improvements \cite{CVPR10-DPM, TPAMI10-DPM, NIPS11-Grammar, PhD-dissert12-DPM}.

Although today's object detectors have far surpassed DPM in detection accuracy, many of them are still deeply influenced by its valuable insights, e.g., mixture models, hard negative mining, bounding box regression, context priming, etc. In 2010, P. Felzenszwalb and R. Girshick were awarded the ``lifetime achievement'' by PASCAL VOC. 

\subsubsection{Milestones: CNN based Two-stage Detectors}

As the performance of hand-crafted features became saturated, the research of object detection reached a plateau after 2010. 
In 2012, the world saw the rebirth of convolutional neural networks \cite{NIPS12-AlexNet}. As a deep convolutional network is able to learn robust and high-level feature representations of an image, a natural question arises: can we introduce it to object detection? R. Girshick \emph{et al.} took the lead to break the deadlocks in 2014 by proposing the Regions with CNN features (RCNN) \cite{CVPR14-RCNN, TPAMI16-RCNN}. Since then, object detection started to evolve at an unprecedented speed. There are two groups of detectors in the deep learning era: ``two-stage detectors'' and ``one-stage detectors'', where the former frames the detection as a ``coarse-to-fine'' process while the latter frames it as to ``complete in one step". 

\begin{figure}[t]
  \centering{\includegraphics[width=\linewidth]{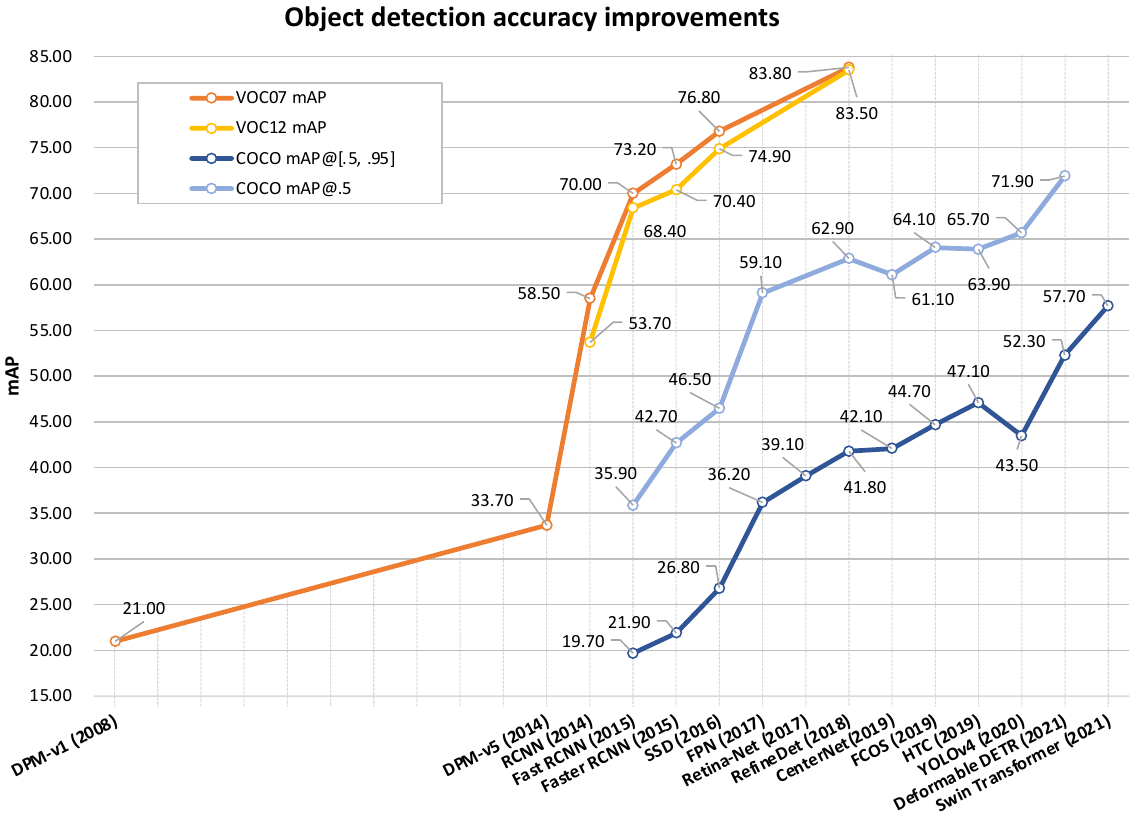}}\\
  \caption{Accuracy improvement of object detection on VOC07, VOC12 and MS-COCO datasets. Detectors in this figure: DPM-v1 \cite{CVPR08-DPM}, DPM-v5 \cite{ECCV14-DPMv5}, RCNN \cite{CVPR14-RCNN}, SPPNet \cite{ECCV14-SPPNet}, Fast RCNN \cite{ICCV15-FastRCNN}, Faster RCNN \cite{NIPS15-FasterRCNN}, SSD \cite{ECCV16-SSD}, FPN \cite{CVPR17-FPN}, Retina-Net \cite{ICCV17-Focal}, RefineDet \cite{CVPR18-SingleShotRefine},  TridentNet \cite{li2019scale} CenterNet \cite{zhou2019objects}, FCOS \cite{tian2019fcos}, HTC \cite{chen2019hybrid}, YOLOv4 \cite{bochkovskiy2020yolov4}, Deformable DETR \cite{zhu2020deformable}, Swin Transformer \cite{liu2021swin}.} \label{figure:accuracy-improvements}
\end{figure}

\textbf{RCNN:}
The idea behind RCNN is simple: It starts with the extraction of a set of object proposals (object candidate boxes) by selective search \cite{ uijlings2013selective}. Then each proposal is rescaled to a fixed size image and fed into a CNN model pretrained on ImageNet (say, AlexNet \cite{NIPS12-AlexNet}) to extract features. Finally, linear SVM classifiers are used to predict the presence of an object within each region and to recognize object categories. RCNN yields a signiﬁcant performance boost on VOC07, with a large improvement of mean Average Precision (mAP) from 33.7\% (DPM-v5 \cite{Online-DPMv5}) to  58.5\%. Although RCNN has made great progress, its drawbacks are obvious: the redundant feature computations on a large number of overlapped proposals (over 2000 boxes from one image) lead to an extremely slow detection speed (14s per image with GPU). Later in the same year, SPPNet \cite{ECCV14-SPPNet} was proposed and has solved this problem. 

\textbf{SPPNet:}
In 2014, K. He \emph{et al.} proposed Spatial Pyramid Pooling Networks (SPPNet) \cite{ECCV14-SPPNet}. Previous CNN models require a fixed-size input, e.g., a 224x224 image for AlexNet \cite{NIPS12-AlexNet}. The main contribution of SPPNet is the introduction of a Spatial Pyramid Pooling (SPP) layer, which enables a CNN to generate a fixed-length representation regardless of the size of the image/region of interest without rescaling it. When using SPPNet for object detection, the feature maps can be computed from the entire image only once, and then fixed-length representations of arbitrary regions can be generated for training the detectors, which avoids repeatedly computing the convolutional features. SPPNet is more than 20 times faster than R-CNN without sacrificing any detection accuracy (VOC07 mAP=59.2\%). Although SPPNet has effectively improved the detection speed, it still has some drawbacks: first, the training is still multi-stage, second, SPPNet only fine-tunes its fully connected layers while simply ignoring all previous layers. Later in the next year, Fast RCNN \cite{ICCV15-FastRCNN} was proposed and solved these problems.

\textbf{Fast RCNN:}
In 2015, R. Girshick proposed Fast RCNN detector \cite{ICCV15-FastRCNN}, which is a further improvement of R-CNN and SPPNet \cite{CVPR14-RCNN, ECCV14-SPPNet}. Fast RCNN enables us to simultaneously train a detector and a bounding box regressor under the same network configurations. On VOC07 dataset, Fast RCNN increased the mAP from 58.5\% (RCNN) to 70.0\% while with a detection speed over 200 times faster than R-CNN. Although Fast-RCNN successfully integrates the advantages of R-CNN and SPPNet, its detection speed is still limited by the proposal detection (see Section \ref{subsec:multi-scale} for more details). Then, a question naturally arises: ``can we generate object proposals with a CNN model?'' Later, Faster R-CNN \cite{NIPS15-FasterRCNN} answered this question.

\textbf{Faster RCNN:}
In 2015, S. Ren \emph{et al.} proposed Faster RCNN detector \cite{NIPS15-FasterRCNN, TPAMI17-FasterRCNN} shortly after the Fast RCNN. Faster RCNN is the first near-realtime deep learning detector (COCO mAP@.5=42.7\%, 
VOC07 mAP=73.2\%,
17fps with ZF-Net \cite{ECCV14-VisCNN}). The main contribution of Faster-RCNN is the introduction of Region Proposal Network (RPN) that enables nearly cost-free region proposals. From R-CNN to Faster RCNN, most individual blocks of an object detection system, e.g., proposal detection, feature extraction, bounding box regression, etc, have been gradually integrated into a unified, end-to-end learning framework. Although Faster RCNN breaks through the speed bottleneck of Fast RCNN, there is still computation redundancy at the subsequent detection stage. Later on, a variety of improvements have been proposed, including RFCN \cite{NIPS16-RFCN} and Light head RCNN \cite{CVPR18-LightHead}. (See more details in Section \ref{sec:speedup}.)

\textbf{Feature Pyramid Networks (FPN):}
In 2017, T.-Y. Lin \emph{et al.} proposed FPN \cite{CVPR17-FPN}. Before FPN, most of the deep learning based detectors run detection only on the feature maps of the networks' top layer. Although the features in deeper layers of a CNN are beneficial for category recognition, it is not conducive to localizing objects. To this end, a top-down architecture with lateral connections is developed in FPN for building high-level semantics at all scales. Since a CNN naturally forms a feature pyramid through its forward propagation, the FPN shows great advances for detecting objects with a wide variety of scales. Using FPN in a basic Faster R-CNN system, it achieves state-of-the-art single model detection results on the COCO dataset without bells and whistles (COCO mAP@.5=59.1\%).
FPN has now become a basic building block of many latest detectors.

\subsubsection{Milestones: CNN based One-stage Detectors}

Most of the two-stage detectors follow a coarse-to-fine processing paradigm. The coarse strives to improve recall ability, while the fine refines the localization on the basis of the coarse detection, and places more emphasis on the discriminate ability. They can easily attain a high precision without any bells and whistles, but rarely employed in engineering due to the poor speed and enormous complexity. On the contrary, one-stage detectors can retrieve all objects in one-step inference. They are well-liked by mobile devices with real-time and easy-deployed features, but their performance suffers noticeably when detecting dense and small objects.

\textbf{You Only Look Once (YOLO):}
YOLO was proposed by R. Joseph \emph{et al.} in 2015. It was the first one-stage detector in the deep learning era \cite{CVPR16-YOLO}. YOLO is extremely fast: a fast version of YOLO runs at 155fps with VOC07 mAP=52.7\%, while its enhanced version runs at 45fps with VOC07 mAP=63.4\%.
YOLO follows a totally different paradigm from two-stage detectors: to apply a single neural network to the full image. This network divides the image into regions and predicts bounding boxes and probabilities for each region simultaneously. In spite of its great improvement of detection speed, YOLO suffers from a drop of localization accuracy compared with two-stage detectors, especially for some small objects. YOLO's subsequent versions \cite{CVPR17-YOLOv2, redmon2018yolov3, bochkovskiy2020yolov4} and the latter proposed SSD \cite{ECCV16-SSD} has paid more attention to this problem. Recently, YOLOv7 \cite{wang2022yolov7}, a follow-up work from YOLOv4 team, has been proposed. It outperforms most existing object detectors in terms of speed and accuracy (range from 5 FPS to 160 FPS) by introducing optimized structures like dynamic label assignment and model structure reparameterization.

\textbf{Single Shot MultiBox Detector (SSD):}
SSD \cite{ECCV16-SSD} was proposed by W. Liu \emph{et al.} in 2015. The main contribution of SSD is the introduction of the multi-reference and multi-resolution detection techniques (to be introduced in Section \ref{subsec:multi-scale}), which significantly improves the detection accuracy of a one-stage detector, especially for some small objects. SSD has advantages in terms of both detection speed and accuracy 
(COCO mAP@.5=46.5\%, a fast version runs at 59fps).
The main difference between SSD and previous detectors is that SSD detects objects of different scales on different layers of the network, while the previous ones only run detection on their top layers.

\textbf{RetinaNet:}
Despite its high speed and simplicity, the one-stage detectors have trailed the accuracy of two-stage detectors for years. T.-Y. Lin \emph{et al.} have explored the reasons behind and proposed RetinaNet in 2017 \cite{ICCV17-Focal}. They found that the extreme foreground-background class imbalance encountered during the training of dense detectors is the central cause. To this end, a new loss function named ``focal loss'' has been introduced in RetinaNet by reshaping the standard cross entropy loss so that detector will put more focus on hard, misclassified examples during training. Focal Loss enables the one-stage detectors to achieve comparable accuracy of two-stage detectors while maintaining a very high detection speed (COCO mAP@.5=59.1\%).

\textbf{CornerNet:}
Previous methods primarily used anchor boxes to provide classification and regression references. Objects frequently exhibit variation in terms of number, location, scale, ratio, etc. They have to follow the path of setting up a large number of reference boxes to better match ground truths in order to achieve high performance. However, the network would suffer from further category imbalance, lots of hand-designed hyper-parameters, and a long convergence time. To address these problems, H. Law \emph{et al} \cite{law2018cornernet} discard the previous detection paradigm, and view the task as a keypoint (corners of a box) prediction problem. After obtaining the key points, it will decouple and re-group the corner points using extra embedding information to form the bounding boxes. CornerNet outperforms most one-stage detectors at that time (COCO mAP@.5=57.8\%).

\textbf{CenterNet:}
X. Zhou \emph{et al} proposed CenterNet \cite{zhou2019objects} in 2019. It also follows a keypoint-based detection paradigm, but eliminates costly post-processes such as group-based keypoint assignment (in CornerNet \cite{law2018cornernet}, ExtremeNet \cite{zhou2019bottom}, etc) and NMS, resulting in a fully end-to-end detection network. CenterNet considers an object to be a single point (the object's center) and regresses all of its attributes (such as size, orientation, location, pose, etc) based on the reference center point. The model is simple and elegant, and it can integrate 3-D object detection, human pose estimation, optical flow learning, depth estimation, and other tasks into a single framework. Despite using such a concise detection concept, CenterNet can also achieve comparative detection results (COCO mAP@.5=61.1\%).

\textbf{DETR:}
In recent years, Transformers have deeply affected the entire field of deep learning, particularly the field of computer vision. Transformers discard the traditional convolution operator in favor of attention-alone calculation in order to overcome the limitations of CNNs and obtain a global-scale receptive field. In 2020, N. Carion \emph{et al} proposed DETR \cite{carion2020end}, where they viewed object detection as a set prediction problem and proposed an end-to-end detection network with Transformers. So far, object detection has entered a new era in which objects can be detected without the use of anchor boxes or anchor points. Later, X. Zhu \emph{et al} proposed Deformable DETR \cite{zhu2020deformable} to address the DETR's long convergence time and limited performance on detecting small objects. It achieves state-of-the-art performance on MSCOCO dataset (COCO mAP@.5=71.9\%).

\begin{figure*}
  \centering{\includegraphics[width=\linewidth]{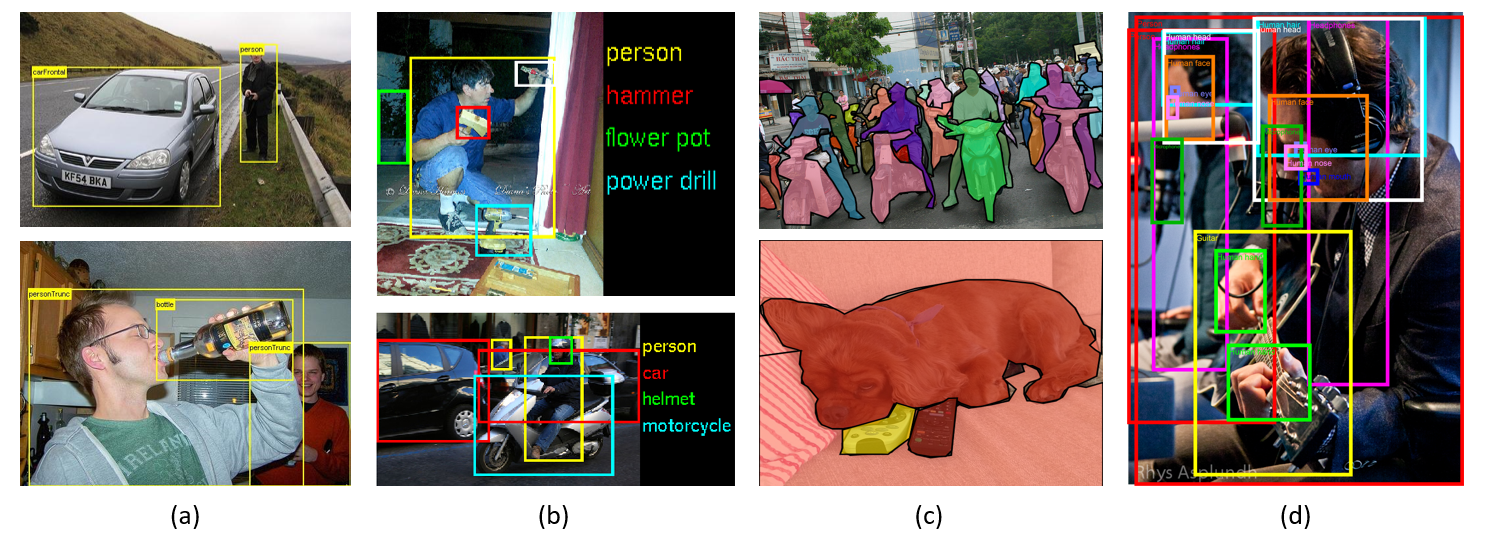}}\\
  \caption{Some example images and annotations in (a) PASCAL-VOC07, (b) ILSVRC, (c) MS-COCO, and (d) Open Images.}\label{figure:dataset-examples}
\end{figure*}

\begin{table*}
\resizebox{\textwidth}{!}{
\centering
    \begin{tabular}{l|rr|rr|rr|rr}
    \toprule
    \multirow{2}{*}{Dataset} &
    \multicolumn{2}{c}{train}  &  \multicolumn{2}{c}{validation} &   \multicolumn{2}{c}{trainval}    & \multicolumn{2}{c}{test} \\
     & images & objects  &  images &objects   &  images &objects  &  images &objects    \\
    \midrule
    VOC-2007 &2,501 &	 6,301 &	 2,510 &	 6,307 &	 5,011 	 & 12,608 &	 4,952 & 14,976 \\
    VOC-2012 & 5,717 & 13,609 & 5,823 & 13,841 & 11,540 & 27,450 & 10,991 & - \\
    ILSVRC-2014 &  456,567 & 478,807 & 20,121 & 55,502 & 476,688 & 534,309 & 40,152  & -  \\
    ILSVRC-2017 &  456,567 & 478,807 & 20,121 & 55,502 & 476,688 & 534,309 & 65,500  & -  \\
    MS-COCO-2015 & 82,783 & 604,907 & 40,504 & 291,875 & 123,287 & 896,782 & 81,434  & - \\
    MS-COCO-2017 &  118,287 & 860,001 & 5,000 & 36,781 & 123,287 & 896,782 & 40,670  & -  \\
    Objects365-2019 & 600,000 & 9,623,000 & 38,000 & 479,000 & 638,000 & 10,102,000& 100,000 & 1,700,00 \\
    OID-2020 &  1,743,042 & 14,610,229  & 41,620 & 303,980 & 1,784,662 & 14,914,209 & 125,436  & 937,327 \\
    \bottomrule
    \end{tabular}}
 \caption{Some well-known object detection datasets and their statistics.}
\label{tab:objdet_datasets}
\end{table*}%

\subsection{Object Detection Datasets and Metrics}\label{subsec:dataset_metirc}

\subsubsection{Datasets} Building larger datasets with less bias is essential for developing advanced detection algorithms. A number of well-known detection datasets have been released in the past 10 years, including the datasets of PASCAL VOC Challenges \cite{IJCV10-VOC, IJCV15-VOC} (e.g., VOC2007, VOC2012), ImageNet Large Scale Visual Recognition Challenge (e.g., ILSVRC2014) \cite{IJCV15-ILSVRC}, MS-COCO Detection Challenge \cite{ECCV14-COCO}, Open Images Dataset \cite{OpenImages,OpenImagesSegmentation}, Objects365 \cite{shao2019objects365}, etc. The statistics of these datasets are given in Table \ref{tab:objdet_datasets}. Fig.\ \ref{figure:dataset-examples} shows some image examples of these datasets. Fig.\ \ref{figure:accuracy-improvements} shows the improvements of detection accuracy on VOC07, VOC12 and MS-COCO datasets from 2008 to 2021.

\textbf{Pascal VOC:}
The PASCAL Visual Object Classes (VOC) Challenges\footnote{\url{http://host.robots.ox.ac.uk/pascal/VOC/}} (from 2005 to 2012) \cite{IJCV10-VOC, IJCV15-VOC} was one of the most important competitions in the early computer vision community. Two versions of Pascal-VOC are mostly used in object detection: VOC07 and VOC12, where the former consists of 5k tr.\ images + 12k annotated objects, and the latter consists of 11k tr.\ images + 27k annotated objects. 20 classes of objects that are common in life are annotated in these two datasets, e.g., ``person'', ``cat'', ``bicycle'', ``sofa'', etc.

\textbf{ILSVRC:}
The ImageNet Large Scale Visual Recognition Challenge (ILSVRC)\footnote{\url{http://image-net.org/challenges/LSVRC/}} \cite{IJCV15-ILSVRC} has pushed forward the state of the art in generic object detection. ILSVRC is organized each year from 2010 to 2017. It contains a detection challenge using ImageNet images \cite{CVPR09-ImageNet}. The ILSVRC detection dataset contains 200 classes of visual objects. The number of its images/object instances is two orders of magnitude larger than VOC.

\textbf{MS-COCO:}
MS-COCO\footnote{\url{http://cocodataset.org/}} \cite{ECCV14-COCO} is one of the most challenging object detection dataset available today. The annual competition based on MS-COCO dataset has been held since 2015. It has less number of object categories than ILSVRC, but more object instances. For example, MS-COCO-17 contains 164k images and 897k annotated objects from 80 categories. Compared with VOC and ILSVRC, the biggest progress of MS-COCO is that apart from the bounding box annotations, each object is further labeled using per-instance segmentation to aid in precise localization. In addition, MS-COCO contains more small objects (whose area is smaller than 1\% of the image) and more densely located objects.
Just like ImageNet in its time, MS-COCO has become the de facto standard for the object detection community.

\textbf{Open Images:}
The year of 2018 sees the introduction of the Open Images Detection (OID) challenge\footnote{\url{https://storage.googleapis.com/openimages/web/index.html}} \cite{Online-OpenImages}, following MS-COCO but at an unprecedented scale. There are two tasks in Open Images: 1) the standard object detection, and 2) the visual relationship detection which detects paired objects in particular relations. For the standard detection task, the dataset consists of 1,910k images with 15,440k annotated bounding boxes on 600 object categories.

\subsubsection{Metrics}

How can we evaluate the accuracy of a detector? This question may have different answers at different times. In the early time's detection research, there are no widely accepted evaluation metrics on detection accuracy. For example, in the early research of pedestrian detection \cite{CVPR05-HOG}, the ``miss rate vs. false positives per window (FPPW)'' was commonly used as the metric. However, the per-window measurement can be flawed and fails to predict full image performance \cite{CVPR09-CaltechDataset}. In 2009, the Caltech pedestrian detection benchmark was introduced \cite{CVPR09-CaltechDataset, TPAMI12-CaltechDataset} and since then, the evaluation metric has changed from FPPW to false positives per-image (FPPI).

In recent years, the most frequently used evaluation for detection is ``Average Precision (AP)'', which was originally introduced in VOC2007. AP is defined as the average detection precision under different recalls, and is usually evaluated in a category-specific manner. The mean AP (mAP) averaged over all categories is usually used as the final metric of performance. To measure the object localization accuracy, the IoU between the predicted box and the ground truth is used to verify whether it is greater than a predefined threshold, say, 0.5. If yes, the object will be identified as ``detected'', otherwise, ``missed''. The 0.5-IoU mAP has then become the de facto metric for object detection.

After 2014, due to the introduction of MS-COCO datasets, researchers started to pay more attention to the accuracy of object localization. Instead of using a fixed IoU threshold, MS-COCO AP is averaged over multiple IoU thresholds between 0.5 and 0.95, which encourages more accurate object localization and may be of great importance for some real-world applications (e.g., imagine there is a robot trying to grasp a spanner).

\begin{figure*}
  \centering{\includegraphics[width=\linewidth]{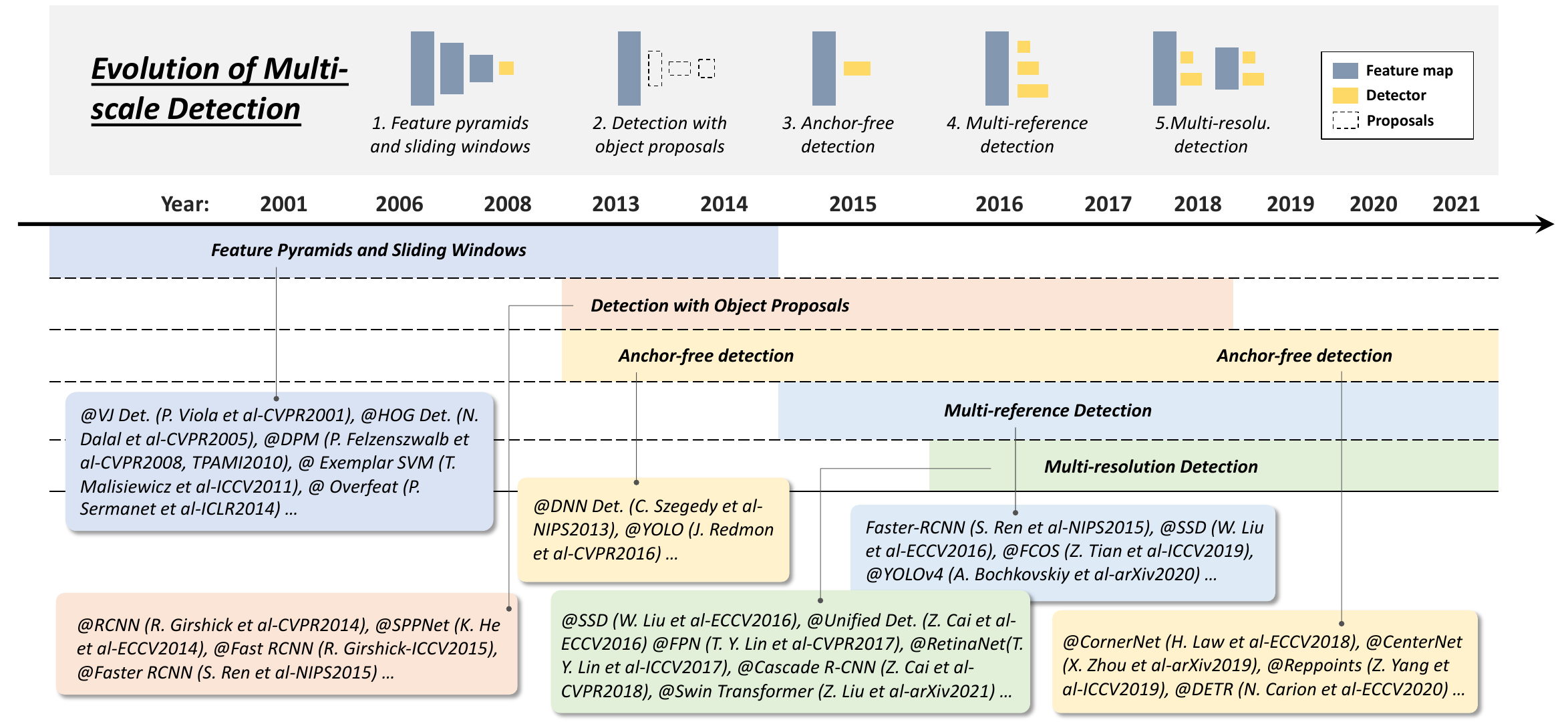}}\\
  \caption{Evolution of multi-scale detection techniques in object detection. Detectors in this figure: VJ Det. \cite{CVPR01-VJ}, HOG Det. \cite{CVPR05-HOG}, DPM \cite{CVPR08-DPM}, Exemplar SVM \cite{ICCV11-Exemplar}, Overfeat \cite{ICLR14-Overfeat}, RCNN \cite{CVPR14-RCNN}, SPPNet \cite{ECCV14-SPPNet}, Fast RCNN \cite{ICCV15-FastRCNN}, Faster RCNN \cite{NIPS15-FasterRCNN}, DNN Det. \cite{NIPS13-DNNDetec}, YOLO \cite{CVPR16-YOLO}, SSD \cite{ECCV16-SSD}, Unified Det. \cite{ECCV16-Unified}, FPN \cite{CVPR17-FPN}, RetinaNet \cite{ICCV17-Focal}, RefineDet \cite{CVPR18-SingleShotRefine}, Cascade R-CNN \cite{cai2018cascade}, Swin Transformer \cite{liu2021swin}, FCOS \cite{tian2019fcos}, YOLOv4 \cite{bochkovskiy2020yolov4}, CornerNet \cite{law2018cornernet}, CenterNet \cite{zhou2019objects}, Reppoints \cite{yang2019reppoints}, DETR \cite{carion2020end}.} \label{figure:evol-multiscale}
\end{figure*}

\subsection{Technical Evolution in Object Detection}

In this section, we will introduce some important building blocks of a detection system and their technical evolutions. First, we describe the multi-scale and context priming on model designing, followed by the sample selection strategy and the design of the loss function in the training process, and lastly, the Non-Maximum Suppression in the inference. The time-stamp in the chart and text is supplied by the publication time of papers. The evolution order shown in the figures is primarily to assist readers in understanding and there may be temporal overlap.

\subsubsection{Technical Evolution of Multi-Scale Detection}\label{subsec:multi-scale}

Multi-scale detection of objects with ``different sizes'' and ``different aspect ratios'' is one of the main technical challenges in object detection. In the past 20 years, multi-scale detection has gone through multiple historical periods, as shown in Fig.\ \ref{figure:evol-multiscale}. 

\textbf{Feature pyramids + sliding windows:}
After the VJ detector, researchers started to pay more attention to a more intuitive way of detection, i.e.\ by building ``feature pyramid + sliding windows''. From 2004, a number of milestone detectors were built based on this paradigm, including the HOG detector, DPM, etc.
They frequently glide a fixed size detection window over the image, paying little attention to "different aspect ratios". To detect objects with a more complex appearance, R. Girshick \emph{et al.} began to seek better solutions outside the feature pyramid. The ``mixture model" \cite{TPAMI10-DPM} was a solution at that time, i.e.\ to train multiple detectors for objects of different aspect ratios. Apart from this, exemplar-based detection \cite{ICCV11-Exemplar, Thesis11-Exemplar} provided another solution by training individual models for every object instance (exemplar).

\textbf{Detection with object proposals:} Object proposals refer to a group of class-agnostic reference boxes that are likely to contain any objects. 
Detection with object proposals helps to avoid the exhaustive sliding window search across an image. We refer readers to the following papers for a comprehensive review on this topic \cite{TPAMI14-ProposalSurvey, ECCV14-ProposalSurvey}. Early time's proposal detection methods followed a bottom-up detection philosophy \cite{CVPR10-WhatIsObj, TPAMI12-WhatIsObj}. 
After 2014, with the popularity of deep CNN in visual recognition, the top-down, learning-based approaches began to show more advantages in this problem \cite{CVPR14-BING, CVPR14-Scalable, NIPS15-FasterRCNN}. Now, the proposal detection gradually slipped out of sight after the rise of one-stage detectors.

\textbf{Deep regression and anchor-free detection:} In recent years, with the increase of GPU's computing power, multi-scale detection has become more and more straightforward and brute-force. The idea of using the deep regression to solve multi-scale problems becomes simple, i.e., to directly predict the coordinates of a bounding box based on the deep learning features \cite{NIPS13-DNNDetec, CVPR16-YOLO}.
After 2018, researchers began to think about the object detection problem from the perspective of keypoint detection. These methods often follow two ideas: One is the group-based method which detects keypoints (corners, centers, or representative points) and then conducts object-wise grouping \cite{law2018cornernet, zhou2019bottom, duan2019centernet, yang2019reppoints}; the other is the group-free method which regards an object as one/many points and then regresses the object attributes (size, ratio, etc.) under the reference of the points \cite{tian2019fcos, zhou2019objects}. 
\begin{figure*}
  \centering{\includegraphics[width=\linewidth]{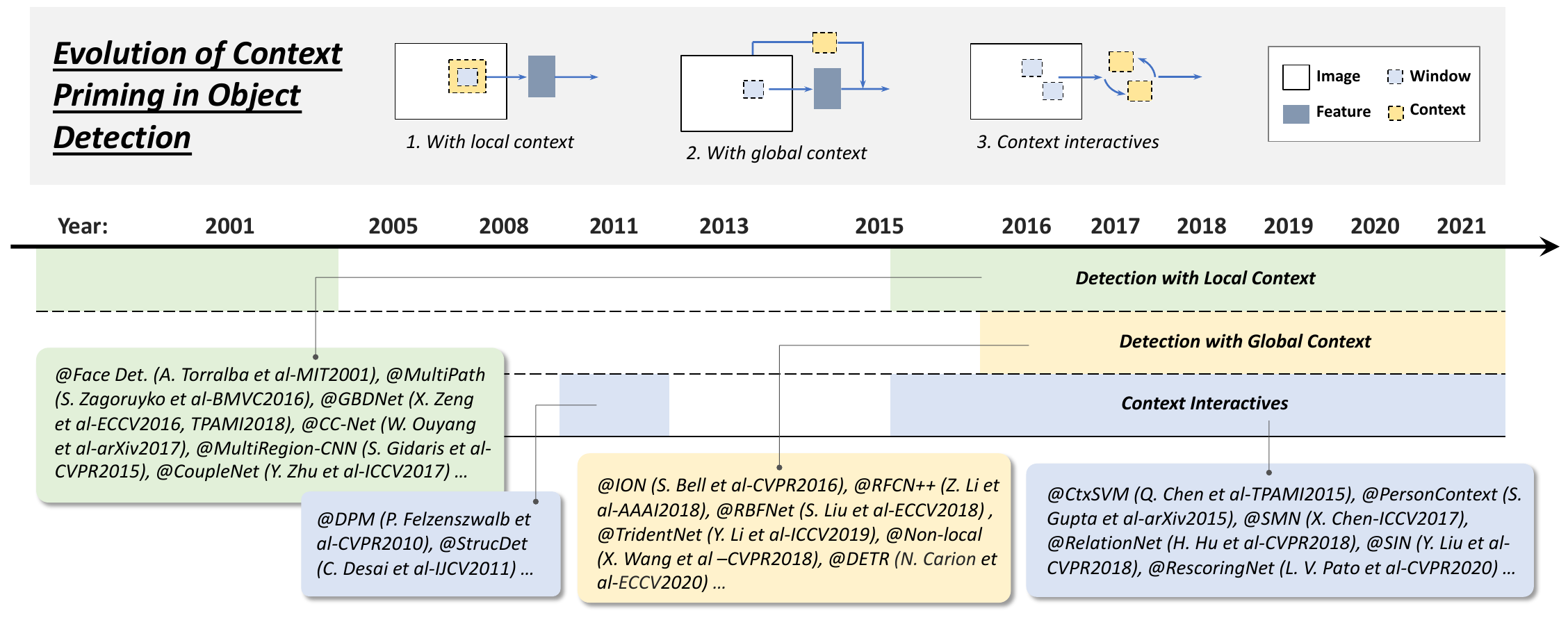}}\\
  \caption{ Evolution of context priming in object detection. Detectors in this figure: Face Det. \cite{MITAILAB01-Faces}, MultiPath \cite{BMVC16-MultiPath}, GBDNet \cite{ECCV16-GBDNet,TPAMI18-GBDNet}, CC-Net \cite{arXiv17-CCNet}, MultiRegion-CNN \cite{CVPR15-MultiRegion}, CoupleNet \cite{ICCV17-CoupleNet}, DPM \cite{CVPR10-DPM,TPAMI10-DPM}, StructDet \cite{IJCV11-Layout}, ION \cite{CVPR16-ION}, RFCN++ \cite{AAAI18-RFCN++}, RBFNet \cite{liu2018receptive}, TridentNet \cite{li2019scale}, Non-local \cite{wang2018non}, DETR \cite{carion2020end}, CtxSVM \cite{TPAMI15-ContextDetCls}, PersonContext \cite{arXiv15-PersonContext}, SMN \cite{chen2017spatial}, RelationNet \cite{hu2018relation}, SIN \cite{CVPR18-StrucInferNet}, RescoringNet \cite{Pato_2020_CVPR}.
  }\label{figure:evol-context}
\end{figure*}

\textbf{Multi-reference/-resolution detection:} Multi-reference detection is now the most used method for multi-scale detection \cite{NIPS15-FasterRCNN, TPAMI17-FasterRCNN, CVPR17-YOLOv2, ECCV16-SSD, bochkovskiy2020yolov4, tian2019fcos}. The main idea of multi-reference detection \cite{NIPS15-FasterRCNN, TPAMI17-FasterRCNN, CVPR17-YOLOv2, ECCV16-SSD, bochkovskiy2020yolov4, tian2019fcos} is to first define a set of references (a.k.a. anchors, including boxes and points) at every location of an image, and then predict the detection box based on these references. Another popular technique is multi-resolution detection \cite{ECCV16-SSD, CVPR17-FPN, ECCV16-Unified, cai2018cascade, liu2021swin}, i.e.\ by detecting objects of different scales at different layers of the network. Multi-reference and multi-resolution detection have now become two basic building blocks in the state-of-the-art object detection systems.

\subsubsection{Technical Evolution of Context Priming}

Visual objects are usually embedded in a typical context with the surrounding environments. Our brain takes advantage of the associations among objects and environments to facilitate visual perception and cognition \cite{CVPR09-EmpContext}. Context priming has long been used to improve detection. Fig.\ \ref{figure:evol-context} shows the evolution of context priming in object detection.

\textbf{Detection with local context:} Local context refers to the visual information in the area that surrounds the object to detect. It has long been acknowledged that local context helps improve object detection. In the early 2000s, Sinha and Torralba \cite{MITAILAB01-Faces} found that the inclusion of local contextual regions such as the facial bounding contour substantially improves face detection performance. Dalal and Triggs also found that incorporating a small amount of background information improves the accuracy of pedestrian detection \cite{CVPR05-HOG}. Recent deep learning based detectors can also be improved with local context by simply enlarging the networks' receptive field or the size of object proposals \cite{BMVC16-MultiPath, ECCV16-GBDNet, TPAMI18-GBDNet, ACCV16-RCNNforSmall, arXiv17-CCNet, CVPR15-MultiRegion, ICCV17-CoupleNet}.

\textbf{Detection with global context:} Global context exploits scene configuration as an additional source of information for object detection. For early time detectors, a common way of integrating global context is to integrate a statistical summary of the elements that comprise the scene, like Gist \cite{CVPR09-EmpContext}. For recent detectors, there are two methods to integrate the global context. The first method is to take advantage of deep convolution, dilated convolution, deformable convolution, pooling operation \cite{li2019scale, liu2018receptive, AAAI18-RFCN++} to receive a large receptive field (even larger than the input image). But now, researchers have explored the potential to apply attention based mechanisms (non-local, transformers, etc.) to achieve a full-image receptive field and have obtained great success \cite{wang2018non, carion2020end}. The second method is to think of the global context as a kind of sequential information and to learn it with the recurrent neural networks \cite{CVPR16-ION, TMM17-AttenContext}.

\textbf{Context interactive:} Context interactive refers to the constraints and dependencies that conveys between visual elements. Some recent researches suggested that modern detectors can be improved by considering context interactives. Some recent improvements can be grouped into two categories, where the first one is to explore the relationship between individual objects \cite{TPAMI10-DPM, TPAMI15-ContextDetCls, IJCV11-Layout, hu2018relation, chen2017spatial, Pato_2020_CVPR}, and the second one is to explore the dependencies between objects and scenes \cite{arXiv15-PersonContext, CVPR18-StrucInferNet}.

\begin{figure*}
  \centering{\includegraphics[width=\linewidth]{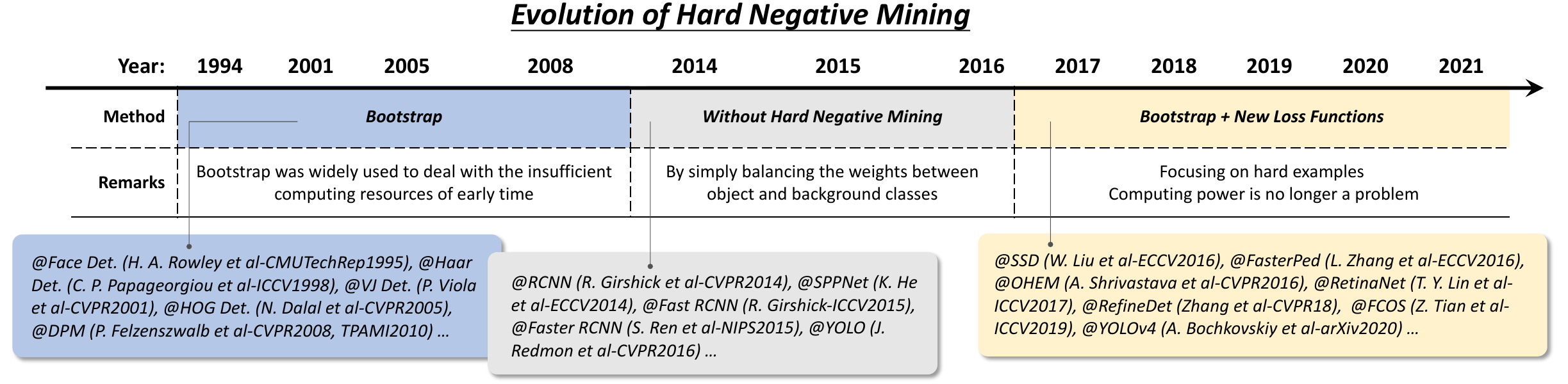}}\\
  \caption{Evolution of hard negative mining techniques in object detection. Detectors in this figure: Face Det. \cite{CMUTechRep95-Face}, Haar Det. \cite{ICCV98-GeneralHaar}, VJ Det. \cite{CVPR01-VJ}, HOG Det. \cite{CVPR05-HOG}, DPM \cite{CVPR08-DPM,TPAMI10-DPM}, RCNN \cite{CVPR14-RCNN}, SPPNet \cite{ECCV14-SPPNet}, Fast RCNN \cite{ICCV15-FastRCNN}, Faster RCNN \cite{NIPS15-FasterRCNN}, YOLO \cite{CVPR16-YOLO}, SSD \cite{ECCV16-SSD}, FasterPed \cite{ECCV16-FasterPed}, OHEM \cite{CVPR16-OHEM}, RetinaNet \cite{ICCV17-Focal}, RefineDet \cite{CVPR18-SingleShotRefine}, FCOS \cite{tian2019fcos}, YOLOv4 \cite{bochkovskiy2020yolov4}.
  }\label{figure:evol-hardnegmining}
\end{figure*}

\subsubsection{Technical Evolution of Hard Negative Mining}

The training of a detector is essentially an imbalanced learning problem. In the case of sliding window based detectors, the imbalance between backgrounds and objects could be as extreme as $10^7{:}1$ \cite{TPAMI14-ProposalSurvey}. In this case, using all backgrounds will be harmful to training as the vast number of easy negatives will overwhelm the learning process. Hard negative mining (HNM) aims to overcome this problem. The technical evolution of HNM is shown in Fig.\ \ref{figure:evol-hardnegmining}.

\textbf{Bootstrap:} Bootstrap in object detection refers to a group of training techniques in which the training starts with a small part of background samples and then iteratively adds new miss-classified samples. In early times detectors, bootstrap was commonly used with the purpose of reducing the training computations over millions of backgrounds \cite{CMUTechRep95-Face, ICCV98-GeneralHaar, CVPR01-VJ}. Later it became a standard technique in DPM and HOG detectors \cite{CVPR05-HOG, CVPR08-DPM} for solving the data imbalance problem. 

\textbf{HNM in deep learning based detectors:} In the deep learning era, due to the increase of computing power, bootstrap was shortly discarded in object detection during 2014-2016 \cite{CVPR14-RCNN, ECCV14-SPPNet, ICCV15-FastRCNN, NIPS15-FasterRCNN, CVPR16-YOLO}. To ease the data-imbalance problem during training, detectors like Faster RCNN and YOLO simply balance the weights between the positive and negative windows. However, researchers later noticed this cannot completely solve the imbalanced problem \cite{ICCV17-Focal}. To this end, the bootstrap was re-introduced to object detection after 2016 \cite{ECCV16-SSD, ECCV16-FasterPed, CVPR16-OHEM, CVPR18-SingleShotRefine}. An alternative improvement is to design new loss functions \cite{ICCV17-Focal} by reshaping the standard cross entropy loss so that it will put more focus on hard, misclassified examples \cite{ICCV17-Focal}.

\subsubsection{Technical Evolution of Loss Function}\label{subsec:lossfunc}
The loss function measures how well the model matches the data (i.e., the deviation of the predictions from the true labels). Calculating the loss yields the gradients of the model weights, which can subsequently be updated by backpropagation to better suit the data. Classification loss and localization loss make up the supervision of the object detection problem, seeing Eq. \ref{eq:loss_sum}. A general form of the loss function can be written as follows:
\begin{equation}
\begin{split}
L(p, p^*, t, t^*) &= L_{cls.}(p, p^*) + \beta I(t)L_{loc.}(t, t^*) \\
I(t) &=
\begin{cases}
1 & \text{IoU}\{a, a^*\}>\eta \\
0 & \text{else}
\end{cases}
\end{split} \label{eq:loss_sum}
\end{equation}
where $t$ and $t^*$ are the locations of predicted and ground-truth bounding boxes, $p$ and $p^*$ are their category probabilities. $\text{IoU}\{a, a^*\}$ is the IoU between the reference box/point $a$ and its ground-truth $a^*$. $\eta$ is an IoU threshold, say, 0.5. If an anchor box/point does not match any objects, its localization loss does not count in the final loss.

\textbf{Classification loss: }
Classification loss is used to evaluate the divergence of the predicted category from the actual category, which was not thoroughly investigated in prevIoUs work such as YOLOv1 \cite{CVPR16-YOLO} and YOLOv2 \cite{CVPR17-YOLOv2} employing MSE/L2 loss (Mean Squared Error). Later, CE loss (Cross-Entropy) is typically used \cite{TPAMI17-FasterRCNN, redmon2018yolov3, ECCV16-SSD}. L2 loss is a measure in Euclidean space, whereas CE loss can measure distribution differences (termed as a form of likelihood). The prediction of classification is a probability, so CE loss is preferable to L2 loss with greater misclassification cost and lower gradient vanishing effect. For improving categorization efficiency, Label Smooth has been proposed to enhance the model generalization ability and solve the overconfidence problem on noise labels \cite{szegedy2016rethinking, muller2019does}, and Focal loss is designed to solve the problem of category imbalance and differences in classification difficulty \cite{ICCV17-Focal}.

\textbf{Localization loss: }
Localization loss is used to optimize position and size deviation. L2 loss is prevalent in early research \cite{CVPR16-YOLO, CVPR17-YOLOv2, CVPR14-RCNN}, but it is highly affected by outliers and prone to gradient explosion. Combining the benefits of L1 loss and L2 loss, the researchers propose Smooth L1 loss \cite{ICCV15-FastRCNN}, as illustrated in the following formula,
\begin{equation}
\text{Smooth}_{L1}(x) = 
\begin{cases}
0.5x^2 & \text{if } |x| < 1 \\
|x| - 0.5 & \text{else}
\end{cases}
\end{equation}
where $x$ denotes the difference between the target and predicted values. When calculating the error, the above losses treat four numbers $(x, y, w, h)$ representing a bounding box as independent variables, however, a correlation exists between them. Moreover, IoU is utilized to determine if the prediction box corresponds to the actual ground truth box in evaluation. Equal Smooth L1 values will have totally different IoU values, hence IoU loss \cite{ACMMM16-UnitBox} is introduced as follows:
\begin{equation}
\text{IoU loss} = - \log (\text{IoU})
\end{equation}

Following that, several algorithms improved IoU loss. G-IoU (Generalized IoU) \cite{rezatofighi2019generalized} improved the case when IoU loss could not optimize the non-overlapping bounding boxes, i.e., $\text{IoU}=0$. According to Distance-IoU \cite{zheng2020distance}, a successful detection regression loss should meet three geometric metrics: overlap area, center point distance, and aspect ratio. So, based on IoU loss and G-IoU loss, DIoU (Distance IoU) is defined as the distance between the center point of the prediction and the ground truth, and CIoU (Complete IoU) \cite{zheng2020distance} considered the aspect ratio difference on the basis of DIoU.

\subsubsection{Technical Evolution of Non-Maximum Suppression}\label{subsec:nms}

\begin{figure*}
  \centering{\includegraphics[width=\linewidth]{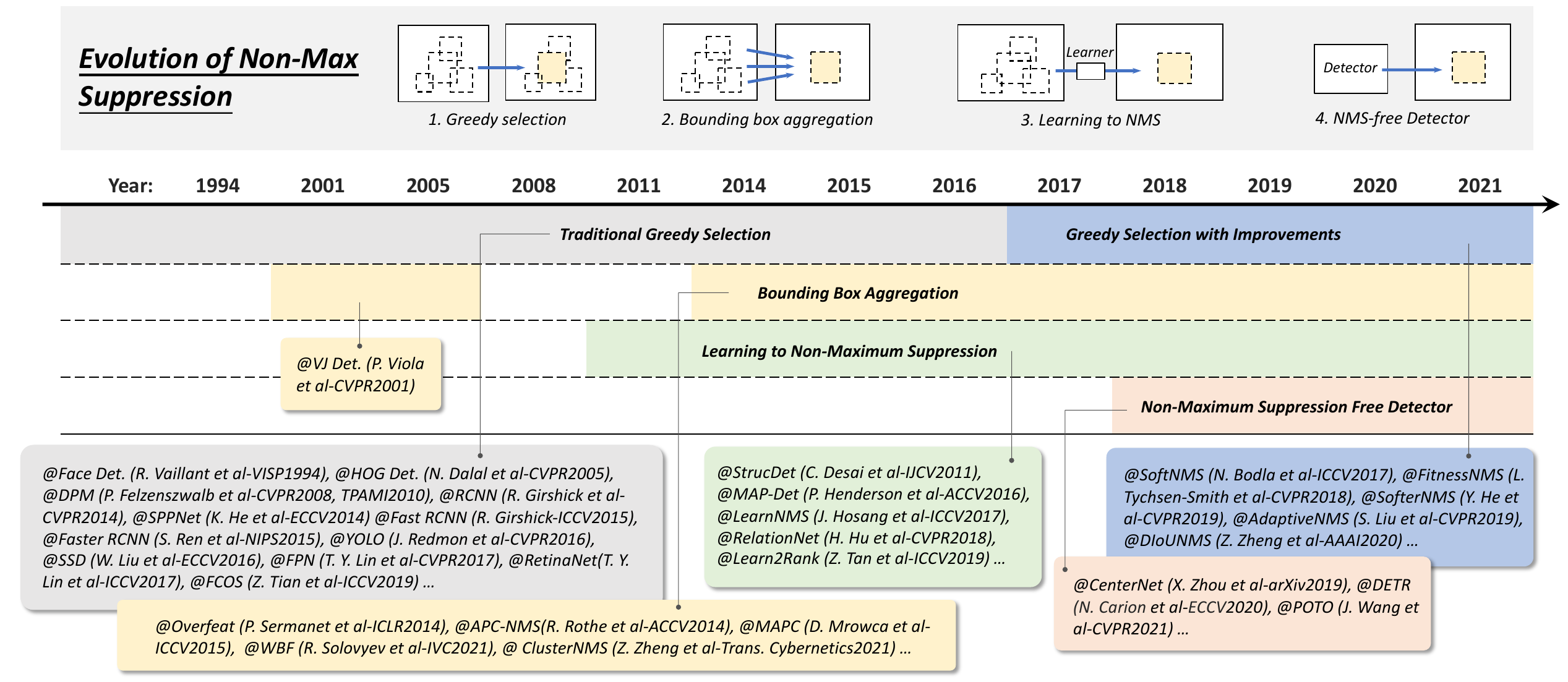}}\\
  \caption{Evolution of non-max suppression (NMS) techniques in object detection from 1994 to 2021: 1) Greedy selection, 2) Bounding box aggregation, 3) Learning to NMS, and 4) NMS-free detection. Detectors in this figure: Face Det. \cite{VISP94-Lecun}, HOG Det. \cite{CVPR05-HOG}, DPM \cite{CVPR08-DPM, TPAMI10-DPM}, RCNN \cite{CVPR14-RCNN}, SPPNet \cite{ECCV14-SPPNet}, Fast RCNN \cite{ICCV15-FastRCNN}, Faster RCNN \cite{NIPS15-FasterRCNN}, YOLO \cite{CVPR16-YOLO}, SSD \cite{ECCV16-SSD}, FPN \cite{CVPR17-FPN}, RetinaNet \cite{ICCV17-Focal}, FCOS \cite{tian2019fcos}, StrucDet \cite{IJCV11-Layout}, MAP-Det \cite{ACCV16-En2EnMAP}, LearnNMS \cite{ICCV17-LearnNMS}, RelationNet \cite{hu2018relation}, Learn2Rank \cite{tan2019learning}, SoftNMS \cite{ICCV17-SoftNMS}, FitnessNMS \cite{CVPR18-FitNMSbdIoU}, SofterNMS \cite{he2019bounding}, AdaptiveNMS \cite{liu2019adaptive}, DIoUNMS \cite{zheng2020distance}, Overfeat \cite{ICLR14-Overfeat}, APC-NMS \cite{ACCV14-NMSMessages}, MAPC \cite{ICCV15-SpatSemReg}, WBF \cite{solovyev2021weighted}, ClusterNMS \cite{zheng2021enhancing}, CenterNet \cite{zhou2019objects}, DETR \cite{carion2020end}, POTO \cite{wang2021end}.
  }\label{figure:evol-nms}
\end{figure*}

As the neighboring windows usually have similar detection scores, the non-maximum suppression is used as a post-processing step to remove the replicated bounding boxes and obtain the final detection result. At early times of object detection, NMS was not always integrated \cite{IJCV00-TrainableHaar}. This is because the desired output of an object detection system was not entirely clear at that time. Fig.\ \ref{figure:evol-nms} shows the evolution of NMS in the past 20 years.

\textbf{Greedy selection:} Greedy selection is an old-fashioned but the most popular way to perform NMS. The idea behind it is simple and intuitive: for a set of overlapped detections, the bounding box with the maximum detection score is selected while its neighboring boxes are removed according to a predefined overlap threshold. Although greedy selection has now become the de facto method for NMS, it still has some space for improvement. First, the top-scoring box may not be the best fit. Second, it may suppress nearby objects. Finally, it does not suppress false positives \cite{ACCV14-NMSMessages}. Many works have been proposed to solve the problems mentioned above \cite{ICCV17-SoftNMS, he2019bounding, liu2019adaptive, zheng2020distance}.

\textbf{Bounding Box aggregation:} BB aggregation is another group of techniques for NMS \cite{CVPR01-VJ, ACCV14-NMSMessages, ICLR14-Overfeat,ICCV15-SpatSemReg} with the idea of combining or clustering multiple overlapped bounding boxes into one final detection. The advantage of this type of method is that it takes full consideration of object relationships and their spatial layout \cite{solovyev2021weighted, zheng2021enhancing}. Some well-known detectors use this method, such as the VJ detector \cite{CVPR01-VJ} and the Overfeat (winner of ILSVRC-13 localization task) \cite{ICLR14-Overfeat}.

\textbf{Learning based NMS:} A recent group of NMS improvements that have recently received much attention is learning based NMS \cite{ICCV17-LearnNMS, ACCV16-En2EnMAP, CVPR15-En2EnDPM, IJCV11-Layout, hu2018relation, tan2019learning}. The main idea is to think of NMS as a filter to re-score all raw detections and to train the NMS as part of a network in an end-to-end fashion or train a net to imitate NMS's behavior. These methods have shown promising results in improving occlusion and dense object detection over traditional hand-crafted NMS methods.

\textbf{NMS-free detector:} To release from NMS and achieve a fully end-to-end object detection training network, researchers developed a series of methods to complete one-to-one label assignment (a.k.a. one object with just one prediction box) \cite{zhou2019objects, carion2020end, wang2021end}. These methods frequently adhere to a rule that calls for the use of the highest-quality box for training in order to achieve free NMS. NMS-free detectors are more similar to the human visual perception system and are also a possible way to the future of object detection.

\section{Speed-Up of Detection}\label{sec:speedup}

\begin{figure}
  \centering{\includegraphics[width=\linewidth]{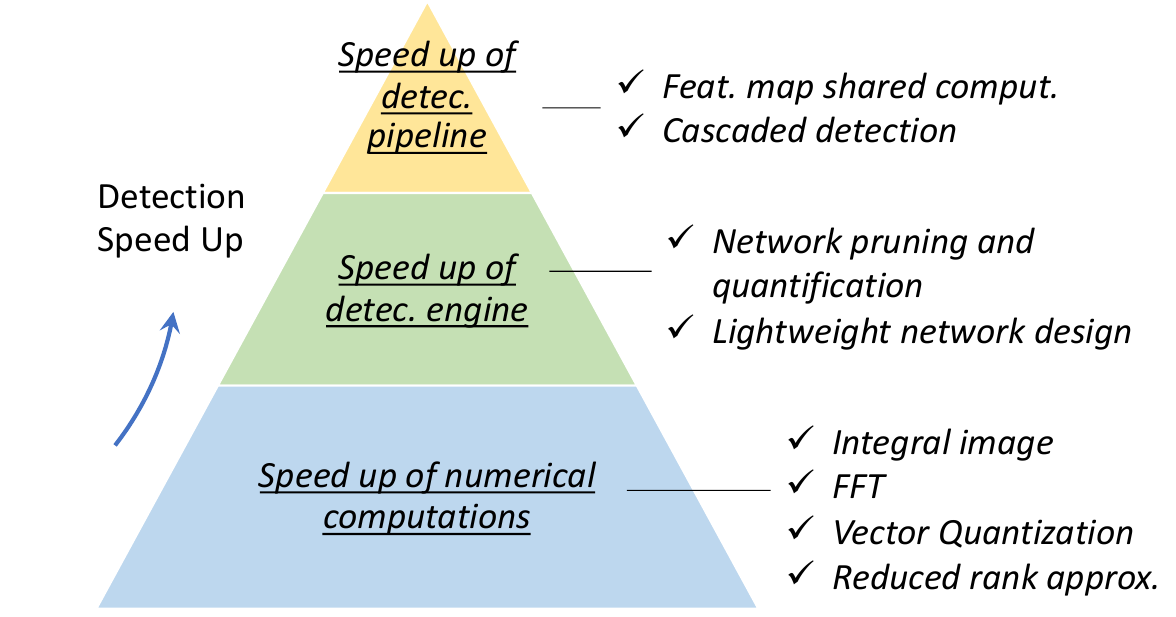}}\\
  \caption{An overview of the speed-up techniques in object detection. }\label{figure:speedup}
\end{figure}

The acceleration of a detector has long been a challenging problem. The speed-up techniques in object detection can be divided into three levels of groups: speed up of ``detection pipeline'', ``detector backbone'', and ``numerical computation''.
, as shown in Fig. \ref{figure:speedup}. Refer to \cite{zou2019object} for a more detailed version.

\subsection{Feature Map Shared Computation}

Among the different computational stages of a detector, feature extraction usually dominates the amount of computation. The most commonly used idea to reduce the feature computational redundancy is to compute the feature map of the whole image only once \cite{CVPR06-FastHOG, ICCV15-FastRCNN, NIPS15-FasterRCNN}, which have achieved tens or even hundreds of times of acceleration.

\subsection{Cascaded Detection}\label{subsec:cascaded_detc}

Cascaded detection is a commonly used technique \cite{CVPR01-VJ, IJCV01-Coarse2Fine}. It takes a coarse to fine detection philosophy: to filter out most of the simple background windows using simple calculations, then to process those more difficult windows with complex ones. In recent years, cascaded detection has been especially applied to those detection tasks of ``small objects in large scenes'', e.g., face detection \cite{CVPR15-CNNCascadeFace, SPL16-JointFaceCascade}, pedestrian detection \cite{CVPR06-FastHOG, ECCV16-FasterPed, ICCV15-CompAwaCasPed}, etc.

\subsection{Network Pruning and Quantification}
``Network pruning'' and ``network quantification'' are two commonly used methods to speed up a CNN model. The former refers to pruning the network structure or weights and the latter refers to reducing their code length. The research of ``network pruning" can be traced back to as early as the 1980s \cite{NIPS89-OptBrainDamage}. The recent network pruning methods usually take an iterative training and pruning process, i.e., to remove only a small group of unimportant weights after each stage of training, and to repeat those operations \cite{UCLR16-Compress}. The recent works on network quantification mainly focus on network binarization, which aims to compress a network by quantifying its activations or weights to binary variables (say, 0/1) so that the floating-point operation is converted to logical operations.

\subsection{Lightweight Network Design}

The last group of methods to speed up a CNN based detector is to directly design lightweight networks. In addition to some general designing principles like ``fewer channels and more layers'' \cite{CVPR15-CNNConstTime}, some other methods have been proposed in recent years \cite{qin2019thundernet, NIPS2018_7466, huang2018yolo, law2019cornernet, yu2021pp}.

\subsubsection{Factorizing Convolutions}

Factorizing convolutions is the most straightforward way to build a lightweight CNN model. There are two groups of factorizing methods. The first group is to factorize a large convolution filter into a set of small ones \cite{CVPR16-IncepV3, AAAI18-RFCN++, CVPR18-LightHead}, as shown in Fig.\ \ref{figure:conv-speedup} (b). For example, one can factorize a 7x7 filter into three 3x3 filters, where they share the same receptive field but the latter one is more efficient. The second group is to factorize convolutions in their channel dimension \cite{arXiv14-ApproxCNN, PAMI15-ApproxCNN}, as shown in Fig.\ \ref{figure:conv-speedup} (c).

\subsubsection{Group Convolution}

Group convolution aims to reduce the number of parameters in a convolution layer by dividing the feature channels into different groups, and then convolve on each group independently \cite{arXiv17-ShuffleNet, CVPR18-CondenseNet}, as shown in Fig.\ \ref{figure:conv-speedup} (d). If we evenly divide the features into $m$ groups, without changing other configurations, the computation will be theoretically reduced to $1/m$ of that before.

\subsubsection{Depth-wise Separable Convolution}

\begin{figure*}
  \centering{\includegraphics[width=\linewidth]{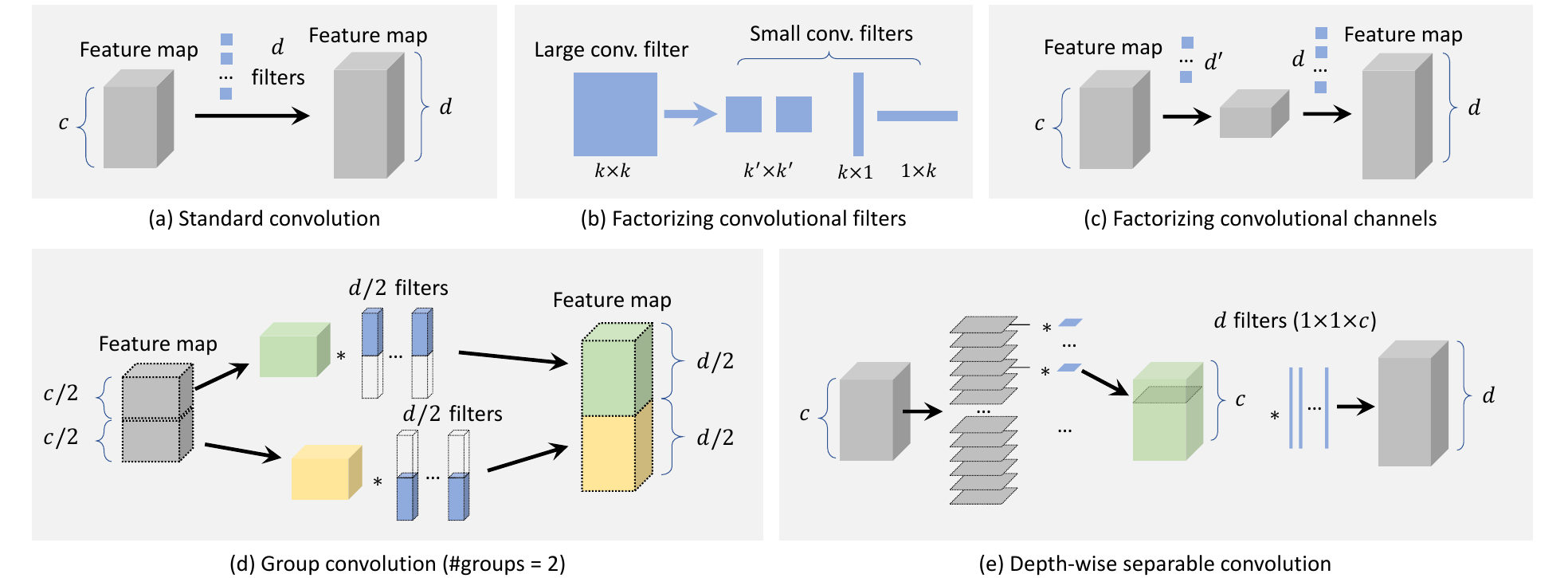}}\\
  \caption{ An overview of speed up methods of a CNN's convolutional layer and the comparison of their computational complexity:  
  (a) Standard convolution: $\mathcal{O}(dk^2c)$.
  (b) Factoring convolutional filters ($k\times k$ $\rightarrow$ $(k^{'}\times k^{'})^2$ or $1\times k, k\times 1$): $\mathcal{O}(dk^{'2}c)$ or $\mathcal{O}(dkc)$.
  (c) Factoring convolutional channels: $\mathcal{O}(d'k^2c) + \mathcal{O}(dk^2d')$.
  (d) Group convolution (\#groups=$m$): $\mathcal{O}(dk^2c/m)$.
  (e) Depth-wise separable convolution: $\mathcal{O}(ck^2) + \mathcal{O}(dc)$.
  }\label{figure:conv-speedup}
\end{figure*}

Depth-wise separable convolution \cite{CVPR17-Xception}, as shown in Fig.\ \ref{figure:conv-speedup} (e) can be viewed as a special case of the group convolution when the number of groups is set equal to the number of channels. Usually, a number of 1x1 filters are used to make a dimension transform so that the final output will have the desired number of channels. By using depth-wise separable convolution, the computation can be reduced from $\mathcal{O}(dk^2c)$ to $\mathcal{O}(ck^2) + \mathcal{O}(dc)$. This idea has been recently applied to object detection and fine-grain classification \cite{CVPR17-MobileNet, CVPR18-MobileNetV2, BMVC18-TinyDSOD}.

\begin{figure*}
  \centering{\includegraphics[width=\linewidth]{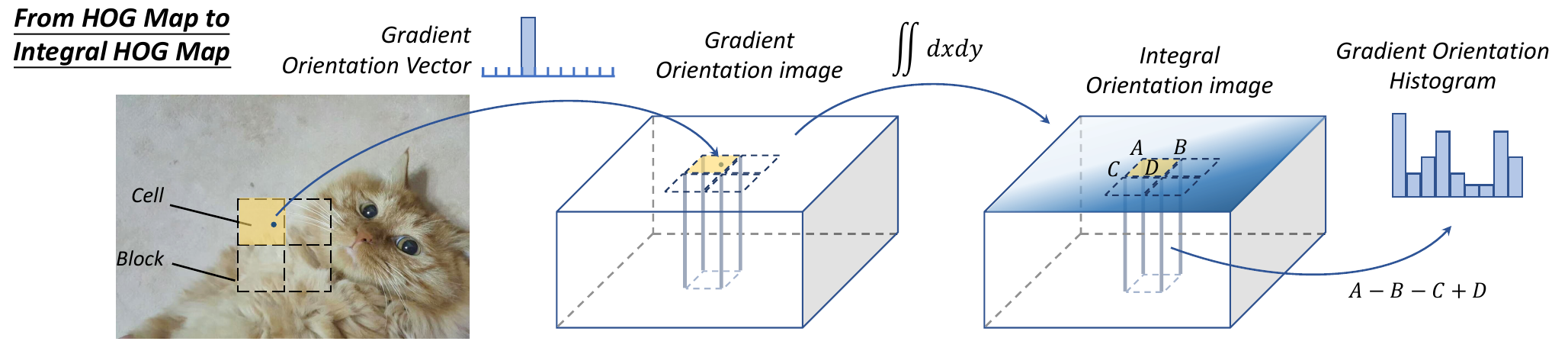}}\\
  \caption{An illustration of how to compute the ``Integral HOG Map'' \cite{CVPR06-FastHOG}. With integral image techniques, we can efficiently compute the histogram feature of any location and any size with constant computational complexity.}\label{figure:integral-HOGmap}
\end{figure*}

\subsubsection{Bottle-neck Design}

A bottleneck layer in a neural network contains few nodes compared to the previous layers. In recent years, the bottle-neck design has been widely used for designing lightweight networks \cite{NIPS2018_7466, ICLR17-Squeezenet, CVPR17-Squeezedet, CVPR18-LightHead, CVPR16-HyperNet}. Among these methods, the input layer of a detector can be compressed to reduce the amount of computation from the very beginning of the detection \cite{NIPS2018_7466, ICLR17-Squeezenet, CVPR17-Squeezedet}. One can also compress the feature map to make it thinner, so that to speed up subsequent detection \cite{CVPR18-LightHead, CVPR16-HyperNet}. 

\subsubsection{Detection with NAS}

Deep learning-based detectors are becoming increasingly sophisticated, relying heavily on hand-crafted network architecture and training parameters. Neural architecture search (NAS) is primarily concerned with defining the proper space of candidate networks, improving strategies for searching quickly and accurately, and validating the searching results at a low cost. When designing a detection model, NAS can reduce the need for human intervention on the design of the network backbone and anchor boxes \cite{chen2019detnas, xu2019auto, ghiasi2019fpn, guo2020hit, wang2020fcos, yao2020sm, jiang2020sp}.

\subsection{Numerical Acceleration}\label{subsec:num_acceleratino}

Numerical Acceleration aims to accelerate object detectors from the bottom of their implementations.

\subsubsection{Speed Up with Integral Image}

The integral image is an important method in image processing. It helps to rapidly calculate summations over image sub-regions. The essence of integral image is the integral-differential separability of convolution in signal processing:
\begin{equation}
f(x) * g(x) = (\int f(x)dx) * (\frac{d g(x)}{dx}),
\end{equation}
where if $d g(x) / dx$ is a sparse signal, then the convolution can be accelerated by the right part of this equation \cite{CVPR01-VJ, NIPS98-Boxlets}. 
The integral image can also be used to speed up more general features in object detection, e.g., color histogram, gradient histogram \cite{ICCV09-HOGLBP, CVPR06-FastHOG, CVPR05-IntegralHist, BMVC09-Integral}, etc. A typical example is to speed up HOG by computing integral HOG maps \cite{ICCV09-HOGLBP, CVPR06-FastHOG}, as shown in Fig.\ \ref{figure:integral-HOGmap}. Integral HOG map has been used in pedestrian detection and has achieved dozens of times' acceleration without losing any accuracy \cite{CVPR06-FastHOG}.

\subsubsection{Speed Up in Frequency Domain}

Convolution is an important type of numerical operation in object detection. As the detection of a linear detector can be viewed as the window-wise inner product between the feature map and detector's weights, which can be implemented by convolutions. The Fourier transform is a very practical way to speed up convolutions, where the theoretical basis is the convolution theorem in signal processing, i.e.\, under suitable conditions, the Fourier transform $F$ of a convolution of two signals $I*W$ is the point-wise product in their Fourier space:
\begin{equation}
I*W = F^{-1}(F(I) \odot F(W))
\end{equation}
where $F$ is Fourier transform, $F^{-1}$ is Inverse Fourier transform, and $\odot$ is the point-wise product. The above calculation can be accelerated by using the Fast Fourier Transform (FFT) and the Inverse FFT (IFFT) \cite{ICLR14-CNNFFT, ECML17-CNNFFT, arXiv14-FBFFT, NIPS15-SpectralCNN}.

\subsubsection{Vector Quantization}

The Vector Quantization (VQ) is a classical quantization method in signal processing that aims to approximate the distribution of a large group of data by a small set of prototype vectors. It can be used for data compression and accelerating the inner product operation in object detection \cite{NIPS13-VQ, ECCV12-VQ}.

\section{Recent Advances in Object Detection}\label{sec:recent-advances}

The continual appearance of new technologies over the past two decades has a considerable influence on object detection, while its fundamental principles and underlying logic have remained unchanged. In the above sections, we introduced the evolution of technology over the past two decades in a large-scale time range to help readers comprehend object detection; in this section, we will focus more on state-of-the-art algorithms in recent years on a short time range to help readers understand object detection. Some are expansions of previously discussed techniques (e.g., Sec. \ref{subsec:beyond} -- \ref{subsec:learning_seg}), while others are novel crossovers that mix concepts (e.g., Sec. \ref{subsec:adversarial} -- \ref{subsec:domain}).

\subsection{Beyond Sliding Window Detection}\label{subsec:beyond}

Since an object in an image can be uniquely determined by its upper left corner and lower right corner of the ground truth box, the detection task, therefore, can be equivalently framed as a pair-wise key points localization problem. One recent implementation of this idea is to predict a heat-map for the corners \cite{law2018cornernet}. Some other methods follow the idea and utilize more key points (corner and center \cite{duan2019centernet}, extreme and center points \cite{zhou2019bottom}, representative points \cite{yang2019reppoints}
) to obtain better performance. Another paradigm views an object as a point/points and directly predicts the object's attributes (e.g. height and width) without grouping. The advantage of this approach is that it can be implemented under a semantic segmentation framework, and there is no need to design multi-scale anchor boxes. Furthermore, by viewing object detection as a set prediction, DETR \cite{carion2020end, zhu2020deformable} completely liberates it in a reference-based framework.

\subsection{Robust Detection of Rotation and Scale Changes} \label{subsec:robust}

In recent years, efforts have been made on robust detection of rotation and scale changes.

\subsubsection{Rotation Robust Detection}\label{subsec:rotation_detec}

Object rotation is common to see in face detection, text detection, and remote sensing object detection. The most straightforward solution to this problem is to perform data augmentation so that an object in any orientation can be well covered by the augmented data distribution \cite{ICIP15-UCASAODDataset}, or to train independent detectors separately for each orientation \cite{GRSL18-Online, JPRS14-COPD}. Designing rotation invariant loss functions is a recent popular solution, where a constraint on the detection loss is added so that the feature of rotated objects keeps unchanged \cite{CVPR16-RIFD, TGRS16-RIFD, cheng2018learning}. Another recent solution is to learn geometric transformations of the objects candidates \cite{CVPR18-RIFaceProgressive, NIPS15-STN, ECCV16-TransformerFace, ding2019learning}. In two-stage detectors, ROI pooling aims to extract a fixed-length feature representation for an object proposal with any location and size. Since the feature pooling usually is performed in Cartesian coordinates, it is not invariant to rotation transform. A recent improvement is to perform ROI pooling in polar coordinates so that the features can be robust to the rotation changes \cite{GRSL18-Online}.

\begin{figure*}
  \centering{\includegraphics[width=\linewidth]{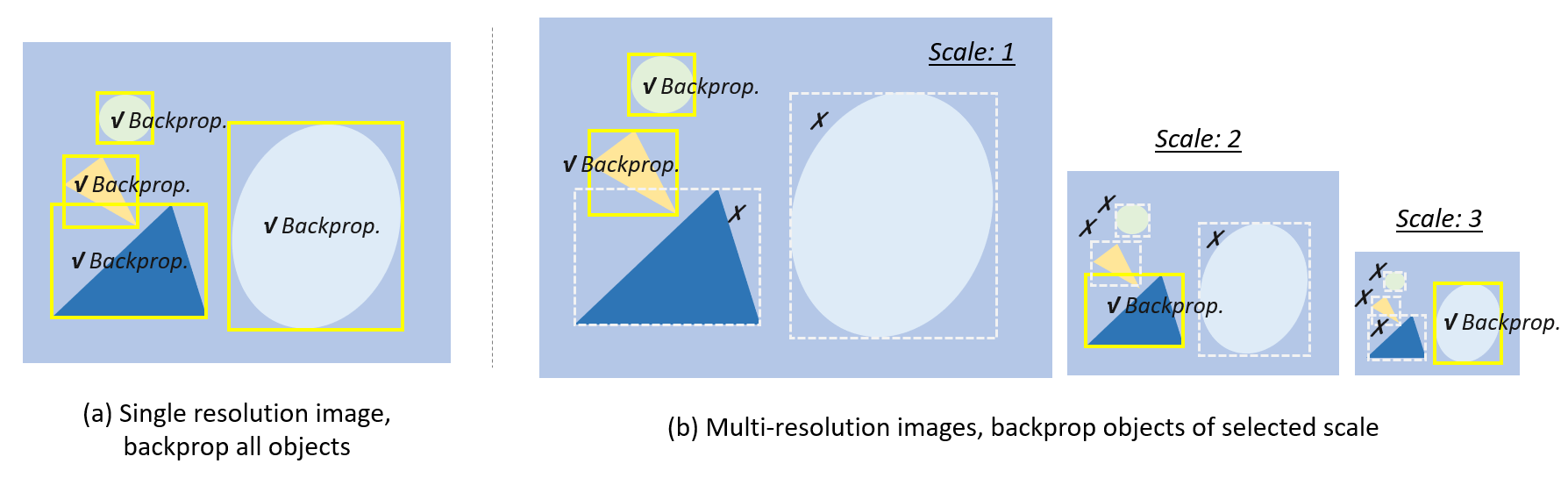}}\\
  \caption{Different training strategies for multi-scale object detection: (a): Training on a single resolution image, back propagate objects of all scales \cite{ECCV14-SPPNet,ICCV15-FastRCNN,NIPS15-FasterRCNN,ECCV16-SSD}. (b) Training on multi-resolution images (image pyramid), back propagate objects of selected scale. If an object is too large or too small, its gradient will be discarded \cite{CVPR18-SNIP,arXiv18-SNIPER,li2019scale}. }\label{figure:snip}
\end{figure*}

\subsubsection{Scale Robust Detection}

Recent studies have been made for scale robust detection at both training and detection stages.

\textbf{Scale adaptive training:} Modern detectors usually re-scale input images to a fixed size and back propagate the loss of the objects in all scales. A drawback of doing this is there will be a ``scale imbalance'' problem. Building an image pyramid during detection could alleviate this problem but not fundamentally \cite{NIPS16-RFCN, CVPR16-ResNet}. A recent improvement is Scale Normalization for Image Pyramids (SNIP) \cite{CVPR18-SNIP}, which builds image pyramids at both training and detection stages and only backpropagates the loss of some selected scales, as shown in Fig.~\ref{figure:snip}. Some researchers have further proposed a more efficient training strategy: SNIP with Efficient Resampling (SNIPER) \cite{arXiv18-SNIPER}, i.e.\ to crop and re-scale an image to a set of sub-regions so that to benefit from large batch training.

\textbf{Scale adaptive detection:} In CNN based detectors, the size of and aspect ratio of anchors are usually carefully designed. A drawback of doing this is the configurations cannot be adaptive to unexpected scale changes. To improve the detection of small objects, some ``adaptive zoom-in'' techniques are proposed in some recent detectors to adaptively enlarge the small objects into the ``larger ones'' \cite{CVPR18-ZoomInNet, CVPR16-ZoomPred}. Another recent improvement is to predict the scale distribution of objects in an image, and then adaptively re-scaling the image according to it \cite{ICCV17-ScaleNet, CVPR17-ScaleHistFace}. 

\subsection{Detection with Better Backbones}\label{subsec:detection}

\begin{figure}[t]
  \centering{\includegraphics[width=\linewidth]{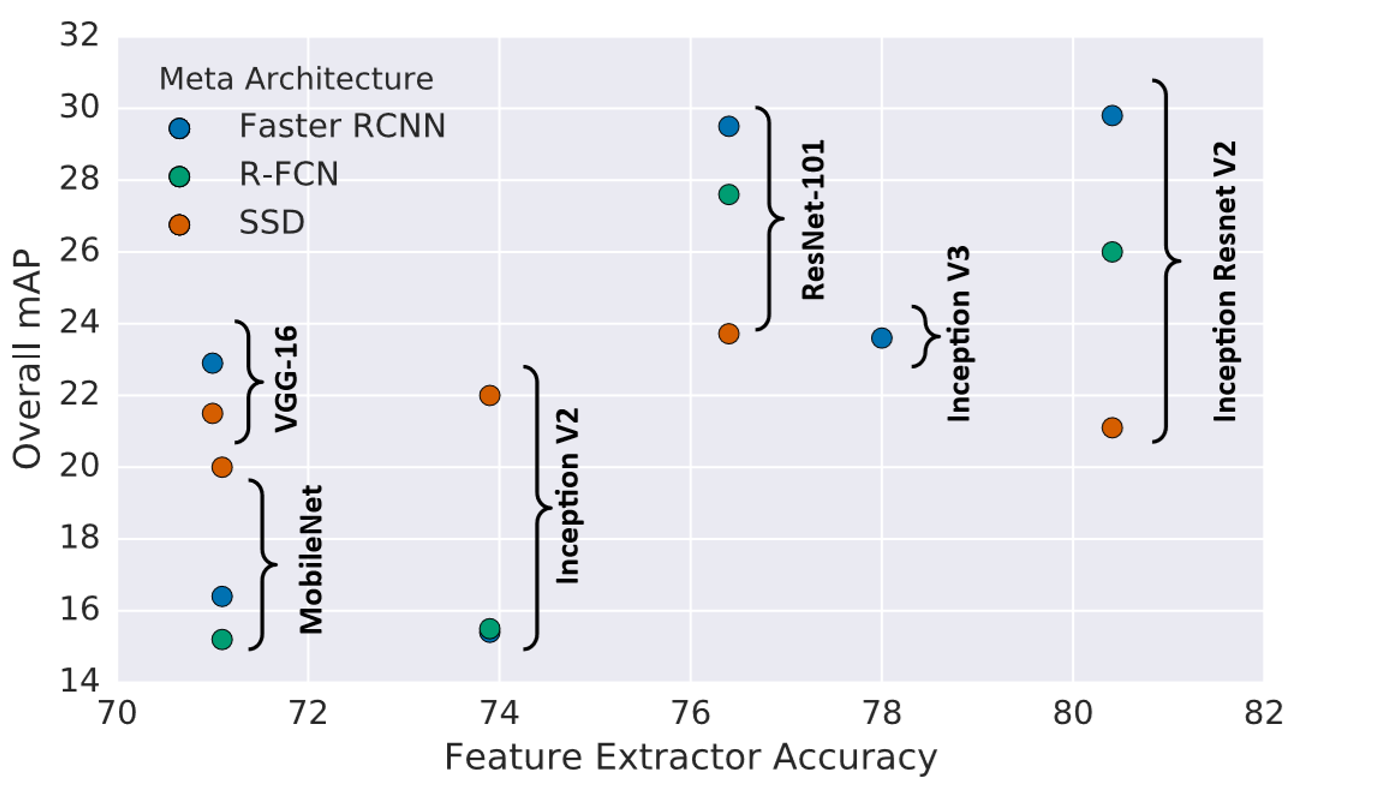}}\\
  \caption{A comparison of detection accuracy of three detectors: Faster RCNN \cite{NIPS15-FasterRCNN}, R-FCN \cite{NIPS16-RFCN} and SSD \cite{ECCV16-SSD} on MS-COCO  dataset with different detection backbones. Image from J. Huang \emph{et al.} CVPR 2017 \cite{CVPR17-SurveyTradeOffs}. }\label{figure:engine-acc}
\end{figure}

The accuracy/speed of a detector depends heavily on the feature extraction networks, a.k.a, backbones, e.g.\ the ResNet \cite{CVPR16-ResNet}, CSPNet \cite{wang2020cspnet}, Hourglass \cite{newell2016stacked}, and Swin Transformer \cite{liu2021swin}. For a detailed introduction of some important detection backbones in deep learning era, we refer readers to the following surveys \cite{arXiv15-SurveyAdvCNN}. Fig.\ \ref{figure:engine-acc} shows the detection accuracy of three well-known detection systems: Faster RCNN \cite{NIPS15-FasterRCNN}, R-FCN \cite{NIPS16-RFCN} and  SSD \cite{ECCV16-SSD} with different backbones \cite{CVPR17-SurveyTradeOffs}. Object detection has recently benefited from the powerful feature extraction capabilities of Transformers. On the COCO dataset, the top-10 detection methods are all transformer-based \footnote{https://paperswithcode.com/sota/object-detection-on-coco}. The performance gap between Transformers and CNNs have been gradually widened.

\subsection{Improvements of Localization}\label{subsec:improve_loc}

To improve localization accuracy, there are two groups of methods in recent detectors: 1) bounding box refinement, and 2) new loss functions for accurate localization.

\subsubsection{Bounding Box Refinement}

The most intuitive way to improve localization accuracy is bounding box refinement, which can be considered as a post-processing of the detection results. One recent method is to iteratively feed the detection results into a BB regressor until the prediction converges to a correct location and size  \cite{arXiv17-CascadeRCNN, ITSC16-RefineNet, IAPR17-RefineFaster}. However, some researchers also claimed that this method does not guarantee the monotonicity of localization accuracy \cite{arXiv17-CascadeRCNN} and may degenerate the localization if the refinement is applied for multiple times.

\subsubsection{New Loss Functions for Accurate Localization}

In most modern detectors, object localization is considered as a coordinate regression problem. However, the drawbacks of this paradigm are obvious. First, the regression loss does not correspond to the final evaluation of localization, especially for some objects with very large aspect ratios. Second, the traditional BB regression method does not provide the confidence of localization. When there are multiple BB's overlapping with each other, this may lead to failure in non-maximum suppression. The above problems can be alleviated by designing new loss functions. The most intuitive improvement is to directly use IoU as the localization loss \cite{ACMMM16-UnitBox, CVPR18-AcqLocConf,rezatofighi2019generalized,zheng2020distance}. Besides, some researchers also tried to improve localization under a probabilistic inference framework \cite{CVPR16-LocNet}. Different from the previous methods that directly predict the box coordinates, this method predicts the probability distribution of a bounding box location.

\subsection{Learning with Segmentation Loss}\label{subsec:learning_seg}

Object detection and semantic segmentation are two fundamental tasks in computer vision. Recent researches suggest object detection can be improved by learning with semantic segmentation losses.

To improve detection with segmentation, the simplest way is to think of the segmentation network as a fixed feature extractor and to integrate it into a detector as auxiliary features \cite{CVPR15-MultiRegion, WACV17-StuffNet, ECCV16-ContexPrimFaster}. The advantage of this approach is that it is easy to implement, while the disadvantage is that the segmentation network may bring additional computation. 

Another way is to introduce an additional segmentation branch on top of the original detector and to train this model with multi-task loss functions (seg.\ + det.) \cite{WACV17-StuffNet,ICCV17-MaskRCNN,chen2019hybrid}. The advantage is the seg.\ brunch will be removed at the inference stage and the detection speed will not be affected. However, the disadvantage is that the training requires pixel-level image annotations.

\subsection{Adversarial Training}\label{subsec:adversarial}

The Generative Adversarial Networks (GAN) \cite{NIPS14-GAN}, introduced by A. Goodfellow \emph{et al.} in 2014, has received great attention in many tasks such as image generation\cite{NIPS14-GAN, radford2015unsupervised}, image style transfer \cite{ICCV17-CycleGAN}, and image super-resolution \cite{CVPR17-GANSuperRes}.

Recently, adversarial training has also been applied to object detection, especially for improving the detection of the small and occluded objects. For small object detection, GAN can be used to enhance the features of small objects by narrowing the representations between small and large ones \cite{CVPR17-GANSmallDetec, ECCV18-SODMTGAN}. To improve the detection of occluded objects, one recent idea is to generate occlusion masks by using adversarial training \cite{CVPR17-AFastRCNN}. Instead of generating examples in pixel space, the adversarial network directly modifies the features to mimic occlusion.

\subsection{Weakly Supervised Object Detection} \label{subsec:wsod}

Training a deep learning based object detector usually requires a large amount of manually labeled data. Weakly Supervised Object Detection (WSOD) aims at easing the reliance on data annotation by training a detector with only image-level annotations instead of bounding boxes \cite{zhang2021weakly}.

Multi-instance learning is a group of supervised learning algorithms that has seen widespread application in WSOD \cite{AI97-MIL, NIPS03-MISVM, TPAMI17-WeaklyMFMIL, CVPR16-NeedNoBB, zhang2020weakly, tang2018pcl, sangineto2018self, zhang2019leveraging}. Instead of learning with a set of instances which are individually labeled, a multi-instance learning model receives a set of labeled bags, each containing many instances. If we consider object candidates in an image as a bag and image-level annotation as the label, then the WSOD can be formulated as a multi-instance learning process.

Class activation mapping is another recent group of methods for WSOD \cite{ICCV17-SoftWeakly, CVPR17-WeaklyCascade}. The research on CNN visualization has shown that the convolutional layer of a CNN behaves as object detectors despite there is no supervision on the location of the object. Class activation mapping shed light on how to enable a CNN with localization capability despite being trained on image-level labels \cite{CVPR16-GlobalPool}.

In addition to the above approaches, some other researchers considered the WSOD as a proposal ranking process by selecting the most informative regions and then training these regions with image-level annotation \cite{CVPR16-WeaklySort}. Some other researchers proposed to mask out different parts of the image. If the detection score drops sharply, then the masked region may contain an object with high probability \cite{WACV16-SelfTaught}. More recently, generative adversarial training has also been used for WSOD \cite{CVPR18-GALWeakly}.

\subsection{Detection with Domain Adaptation}\label{subsec:domain}

The training process of most object detectors can be essentially viewed as a likelihood estimation process under the assumption of independent and identically distributed (i.i.d.) data. Object detection with non-i.i.d. data, especially for some real-world applications, still remains a challenge. Aside from collecting more data or applying proper data augmentation, domain adaptation offers the possibility of narrowing the gap between domains. To obtain domain-invariant feature representation, feature regularization and adversarial training based methods have been explored at the image, category, or object levels \cite{chen2018domain, wang2021domain, hou2021informative, zhu2019adapting, saito2019strong, xu2020exploring}. Cycle-consistent transformation \cite{zhu2017unpaired} has also been applied to bridge the gap between source and target domain \cite{kim2019diversify, inoue2018cross}. Some other methods also incorporate both ideas \cite{hsu2020progressive} to acquire better performance.

\section{Conclusion and Future Directions}\label{sec:conclusion}

Remarkable achievements have been made in object detection over the past 20 years. This paper extensively reviews some milestone detectors, key technologies, speed-up methods, datasets, and metrics in its 20 years of history. Some promising future directions may include but are not limited to the following aspects to help readers get more insights beyond the scheme mentioned above.

\textbf{Lightweight object detection}: Lightweight object detection aims to speed up the detection inference to run on low-power edge devices. Some important applications include mobile augmented reality, automatic driving, smart city, smart cameras, face verification, etc. Although a great effort has been made in recent years, the speed gap between a machine and human eyes still remains large, especially for detecting some small objects or detecting with multi-source information \cite{bosquet2021stdnet, yang2022querydet}.

\textbf{End-to-End object detection}: Although some methods have been developed to detect objects in a fully end-to-end manner (image to box in a network) using one-to-one label assignment training, the majority still use a one-to-many label assignment method where the non-maximum suppression operation is separately designed. Future research on this topic may focus on designing end-to-end pipelines that maintain both high detection accuracy and efficiency \cite{sun2021makes}.

\textbf{Small object detection}: Detecting small objects in large scenes has long been a challenge. Some potential application of this research direction includes counting the population of people in crowd or animals in the open air and detecting military targets from satellite images. 
Some further directions may include the integration of the visual attention mechanisms and the design of high resolution lightweight networks \cite{zhou2021intelligent, cheng2022towards}.

\textbf{3D object detection}: Despite recent advances in 2-D object detection, applications like autonomous driving rely on access to the objects' location and pose in a 3D world. The future of object detection will receive more attention in the 3D world and the utilization of multi-source and multi-view data (e.g., RGB images and 3D lidar points from multiple sensors) \cite{wang2022detr3d, wang2022bridged}.

\textbf{Detection in videos}: Real-time object detection/tracking in HD videos is of great importance for video surveillance and autonomous driving. Traditional object detectors are usually designed under for image-wise detection, while simply ignores the correlations between videos frames. Improving detection by exploring the spatial and temporal correlation under the calculation limitation is an important research direction \cite{cheng2022implicit, zhou2022transvod}.

\textbf{Cross-modality detection}: Object detection with multiple sources/modalities of data, e.g., RGB-D image, lidar, flow, sound, text, video, etc, is of great importance for a more accurate detection system which performs like human-being's perception. Some open questions include: how to immigrate well-trained detectors to different modalities of data, how to make information fusion to improve detection, etc \cite{cong2022cir, wang2022cross}.

\textbf{Towards open-world detection}: Out-of-domain generalization, zero-shot detection, and incremental detection are emerging topics in object detection. The majority of them devised ways to reduce catastrophic forgetting or utilized supplemental information. Humans have an instinct to discover objects of unknown categories in the environment. When the corresponding knowledge (label) is given, humans will learn new knowledge from it, and get to keep the patterns. However, it is difficult for current object detection algorithms to grasp the detection ability of unknown classes of objects. Object detection in the open world aims at discovering unknown categories of objects when supervision signals are not explicitly given or partially given, which holds great promise in applications such as robotics and autonomous driving \cite{feng2022promptdet, zhong2022regionclip}.

Standing on the highway of technical evolutions, we believe this paper will help readers to build a complete road map of object detection and to find future directions of this fast-moving research field.

\bibliographystyle{IEEEtran}
\bibliography{IEEEabrv,egbib}

\end{document}